# Super Neurons

Serkan Kiranyaz[1] *Senior IEEE*, Junaid Malik[2], Mehmet Yamac[3], Mert Duman[2], Ilke Adalioglu[2], Esin Guldogan[5], Turker Ince[4], and Moncef Gabbouj[2] *Fellow IEEE*.

[1] Electrical Engineering, College of Engineering, Qatar University, Qatar; e-mail: mkiranyaz@qu.edu.qa

[2] Department of Computing Sciences, Tampere University, Finland; e-mail: Moncef.gabbouj@tuni.fi, mert.duman@tuni.fi, ilke.adalioglu@tuni.fi

[3] Huawei Technologies Oy (Finland); e-mail: mehmet.yamac@huawei.com

[4] Electrical & Electronics Engineering Department, Izmir University of Economics, Turkey; e-mail: turker.ince@ieu.edu.tr

[5] Microsoft (Finland); e-mail: esinguldogan@microsoft.com

*Abstract*—Self-Organized Operational Neural Networks (Self-ONNs) have recently been proposed as new-generation neural network models with nonlinear learning units, i.e., the generative neurons that yield an elegant level of diversity; however, like its predecessor, conventional Convolutional Neural Networks (CNNs), they still have a common drawback: *localized* (fixed) kernel operations. This severely limits the receptive field and information flow between layers and thus brings the necessity for deep and complex models. It is highly desired to improve the receptive field size without increasing the kernel dimensions. This requires a significant upgrade over the generative neurons to achieve the "non-localized kernel operations" for each connection between consecutive layers. In this article, we present superior (generative) neuron models (or super neurons in short) that allow random or learnable kernel shifts and thus can increase the receptive field size of each connection. The kernel localization process varies among the two super-neuron models. The first model assumes *randomly localized* kernels within a range and the second one learns (optimizes) the kernel locations during training. An extensive set of comparative evaluations against conventional and *deformable* convolutional, along with the generative neurons demonstrates that super neurons can empower Self-ONNs to achieve a superior learning and generalization capability with a minimal computational complexity burden.

## I. Introduction

Generalized Operational Perceptrons (GOPs) [1]-[5] have been proposed as an advanced model of biological neurons with varying nonlinear synaptic connections. Thanks to such a diverse neuron model, GOPs have achieved a superior learning capability on many challenging problems surpassing conventional Multi-Layer Perceptrons (MLPs) and even Extreme Learning Machines (ELMs) with a significant performance gap [1]-[5]. Following the GOPs' main philosophy, Operational Neural Networks (ONNs) [6]-[9] have outperformed CNNs significantly, and achieved a notable learning performance, even on those problems where CNNs entirely fail. Yet**,** ONNs have the following limitations: 1) strict dependability to the operators in the operator set library, 2) requiring a prior search for the best operator set for each layer/neuron which can be highly time-consuming. Self-organized ONNs (Self-ONNs) [10]-[21] have recently been proposed to address these drawbacks with the generative neuron model, which can optimize the nodal operators of each kernel element. Such a capability indeed yields an ultimate neuron heterogeneity that is far superior to what conventional ONNs can offer. Generative neurons can, therefore, replace the traditional "weight optimization" of convolutional neurons with the "nodal function optimization" process. However, their kernels are still "localized" or static, and hence each neuron's receptive field size is determined by its kernel size, and this severely limits the amount of information acquired from the previous layer with such limited and localized kernels. Obviously, using a larger kernel size may be a solution for this; however, it will not only create an increasing complexity issue, but it is also not feasible to determine the optimal kernel size for each connection of the neuron. The aim, therefore, should be to improve the receptive field of each kernel connection by allowing each kernel location to vary while keeping the kernel size the same. Moreover, it would be more beneficial "to learn" or to optimize each kernel location for each connection to the feature maps in the previous layer.

The most prominent approach ever proposed to improve the receptive field size was deformable CNNs [22], [23]. However, the improvements over the regular convolutions were limited or simply none because the kernels of each layer have to be deformed in the same way. Therefore, it is rather a "relocation" operation over each kernel element rather than improving the receptive field size. Furthermore, deformable convolutions further increase the network complexity (number of parameters) and especially the memory overhead significantly. This is why deformable neurons are usually used in only one or a few layers of a (deep) CNN.

To address the aforementioned limitations and drawbacks, the novel and significant contributions of this study can be summarized as follows:

- To accomplish the aim of improving the receptive field size with varying kernel locations, and even optimizing each kernel location, this study proposes a superior generative neuron model (i.e., super neurons in short) with non-localized kernel operations for Self-ONNs.

- This study proposes two super neuron models, each of which has a different kernel localization process: i) random localization within a bias range set for each layer, ii) BP-optimized locations of each kernel.
- Particularly in the latter model, the "what" operator should be used and "where" it should be located, are simultaneously optimized during the BP training. This can be more advantageous for some particular problems where certain optimal kernel locations may exist, or some kernel location topology (or distribution) may be more desirable.
- This study will reveal the *pros* and *cons* of both super neuron models when compared against the generative and convolutional neurons over several challenging problems.
- This study presents a "Proof-of-Concept" experiment where a single (hidden) super neuron will suffice to learn and regress any shifted image from its original. Such a regression can otherwise be performed only by deep CNN models with a high number of neurons.
- Finally, an extensive set of experiments reveals that Self-ONNs with super neurons can outperform equivalent or significantly deeper CNNs in many challenging problems.

The rest of the paper is organized as follows: Section II will briefly present Self-ONNs with generative neurons while the details of the BP training are presented in Appendix A. Section III presents the two super neuron models with non-localized kernel operations in detail and formulates the forward-propagation (FP) and back-propagation (BP). Comparative evaluations among Self-ONNs with generative and super neurons and CNNs over challenging problems are presented in both Section IV and Appendix C. The computational complexity analysis of these networks for both FP and BP is also presented in Section IV. Finally, Section V concludes the paper and suggests topics for future research.

## II. SELF-ORGANIZED OPERATIONAL NEURAL NETWORKS

In biological neurons, during the learning process, the neurochemical characteristics and connection strengths of the synaptic connections are altered, giving rise to new connections and modifying the existing ones. Inspired by this a generative-neuron model for Self-ONNs is formed where each kernel can have a distinct *nonlinear* nodal-operator generated (optimized) during training without any restrictions. As a result, each kernel element of each generative neuron can "customize" its nodal operator to maximize the learning performance. To exemplify this, the nodal operators of the 3x3 kernels the convolutional (CNN), operational (ONN), and generative (Self-ONN) neurons are illustrated in Figure 1. Both convolutional and operational neurons have *static* (fixed) nodal operators (linear and harmonic, respectively) while the generative neuron has *any* arbitrary nodal function, $\Psi$, (including possibly standard functions such as linear and harmonic functions) for each kernel element of each connection.

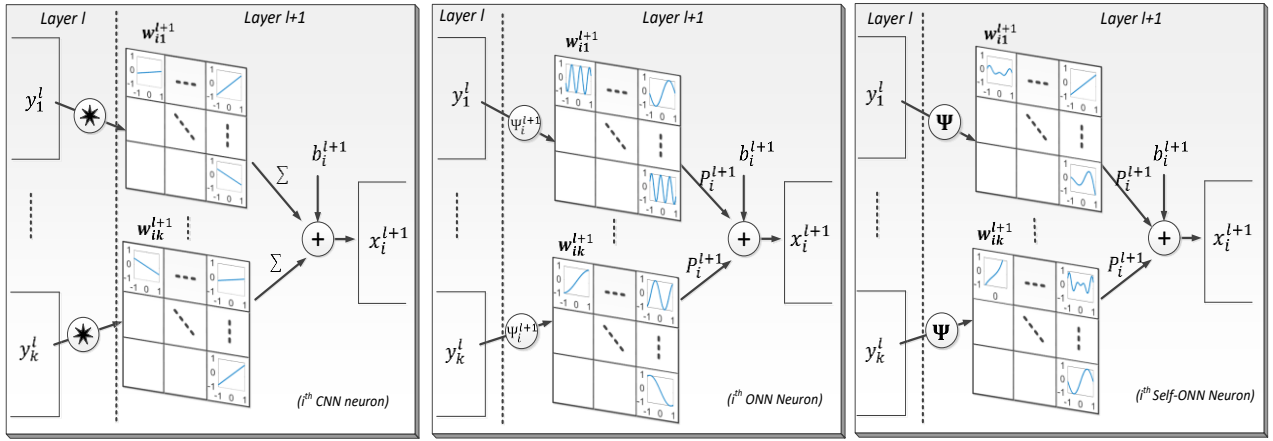

**Figure 1: An illustration of the nodal operations in the kernels of the $i^{th}$ CNN (left), ONN (middle), and Self-ONN (right) neurons at layer $l$+1 [10].**

As illustrated in Figure 1 (middle), for conventional ONNs the input map of the $i^{th}$ neuron at the layer $l$+1, $x_i^{l+1}$, is composed as follows:

$$x_i^{l+1} = b_i^{l+1} + \sum_{k=1}^{N_l} \boldsymbol{oper2D}(y_k^l, w_{ik}^{l+1}, 'NoZeroPad')$$

$$x_i^{l+1}(m,n)\Big|_{(0,0)}^{(M-1,N-1)} = b_i^{l+1} + \sum_{i=1}^{N_{l-1}} \left( P_i^{l+1} \begin{bmatrix} \Psi_i^{l+1}\left(y_k^l(m,n), w_{ik}^{l+1}(0,0)\right), \dots, \\ \Psi_i^{l+1}(y_k^l(m+r,n+t), w_{ik}^{l+1}(r,t)), \dots \end{bmatrix} \right) \quad (1)$$

where $y_k^l$ are the final output maps of the previous layer neurons *operated* with the corresponding kernels, $w_{ik}^{l+1}$, with a particular nodal function, $\Psi_i^{l+1}$ such as linear (multiplication), sinusoid, exponential, Gaussian, chirp, Hermitian, etc. A close look at Eq. (1) reveals the fact that when the pool operator is a summation, i.e. $P_i^{l+1} = \Sigma$, and the nodal operator is a linear function, $\Psi_i^{l+1}(y_k^l(m+r,n+t), w_{ik}^{l+1}(r,t)) = y_k^l(m+r,n+t) \times w_{ik}^{l+1}(r,t)$, for *all* neurons, then the resulting homogenous ONN will be identical to a CNN. Hence, ONNs are a superset of CNNs as the GOPs are a superset of Multi-layer Perceptrons (MLPs). Self-ONNs differ from ONNs by the following two points:

1) each "fixed-in-advance" nodal operator function with a scalar kernel element, $\Psi_i^{l+1}(y_k^l(m+r, n+t), w_{ik}^{l+1}(r,t))$, is *approximated* by the composite nodal operator, $\Psi(y_k^l(m+r, n+t), \boldsymbol{w_{ik}^{l+1}}(r,t))$, as expressed by the Maclaurin series,

2) the *scalar* kernel parameter, $w_{ik}^{l+1}(r,t)$, of the kernel of an ONN neuron, is replaced by a $Q$-dimensional array, $\boldsymbol{w_{ik}^l}(r,t)$.

In this way, any nodal operator function can be approximated by the Maclaurin series near the origin as follows:

$$\Psi(y, \boldsymbol{w_{ik}^{l+1}}(r,t)) = w_{ik}^{l+1}(r,t,0) + w_{ik}^{l+1}(r,t,1)y + w_{ik}^{l+1}(r,t,2)y^2 + \cdots + w_{ik}^{l+1}(r,t,Q)y^Q \quad (2)$$

where $w_{ik}^{l+1}(r,t,q) = \frac{f^{(q)}(0)}{q!}$ is the $q^{th}$ coefficient of the $Q$-order polynomial. During BP training, each $\boldsymbol{w_{ik}^{l+1}}(r,t,q)$ is optimized for the learning problem at hand. Thanks to this ability, there is no need for any operator search for Self-ONNs and arbitrary nodal operators can be customized by the training process as illustrated in Figure 1 (right). This results in enhanced flexibility and diversity over an ONN neuron where only a standard nodal operator function has to be used for all kernels, each connected to an output map of a neuron in the previous layer.

**Table 1: Formula abbreviations and descriptions.**

| Symbol | Description |
|---|---|
| $Q$ | The order of the Maclaurin polynomial |
| Kx × Ky | The size of a kernel, i.e., $\boldsymbol{w_{ik}^{l+1}}$ |
| $\boldsymbol{w_{ik}^{l+1}}(r,t)$ | $Q$-dimensional array of the kernel element (r,t) from the $i^{th}$ neuron in layer $l+1$ to $k^{th}$ neuron in layer $l$. |
| $w_{ik}^{l+1}(r,t,q)$ | The $q^{th}$ element of $\boldsymbol{w_{ik}^l}(r,t)$. |
| $[\alpha_k^i, \beta_k^i] \in \mathbb{Z}[\pm\boldsymbol{\Gamma}]$ | The integer bias pair in x- and y-directions within the range limit, $\boldsymbol{\Gamma}$ for the $i^{th}$ neuron in the current layer connected to the $k^{th}$ neuron in the previous layer. |
| $(\alpha_k^i, \beta_k^i) \in \mathbb{R}[\pm\boldsymbol{\Gamma}]$ | The real bias pair in x- and y-directions within the range limit, $\boldsymbol{\Gamma}$ for the $i^{th}$ neuron in the current layer connected to the $k^{th}$ neuron in the previous layer. |
| $\zeta\alpha_k^i, \zeta\beta_k^i$ | The fractional components of $(\alpha_k^i, \beta_k^i) \in \mathbb{R}[\pm\boldsymbol{\Gamma}]$ where $\zeta\alpha_k^i = \alpha_k^i - \lfloor\alpha_k^i\rfloor$ and $\zeta\beta_k^i = \beta_k^i - \lfloor\beta_k^i\rfloor$. |
| $(y_k^l)$ | The 2D output feature map of the $k^{th}$ neuron at the $l^{th}$ layer. |
| $\Psi(y_k^l, \boldsymbol{w_{ik}^{l+1}}(r,t))$ | The nodal operator function approximated the Maclaurin series over the elements (coefficients) of $\boldsymbol{w_{ik}^l}(r,t)$. |
| $P_i^{l+1} = \Sigma$ | The pooling operator is fixed to summation for this study. |
| $\boldsymbol{T}^{(\alpha_k^i, \beta_k^i)}(y_k^l)$ | The shift operator for $y_k^l$ by the bias, $[\alpha_k^i, \beta_k^i]$. |
| $\vec{y}_k^l = \boldsymbol{T}^{(\alpha_k^i, \beta_k^i)}(y_k^l)$ | The shifted version of the output map $y_k^l$ by the bias, $[\alpha_k^i, \beta_k^i]$. |
| $\Delta_i^{l+1}$ | Given the cost function, $E$, the 2D delta error map of the input feature map, $x_i^{l+1}(m,n)$ at layer $l+1$. Specifically, $\Delta_i^{l+1}(m,n) = \frac{\partial E}{\partial x_i^{l+1}(m,n)}$ |
| $\Delta y_k^l$ | Given the cost function, $E$, the 2D sensitivity of the output map, $y_k^l(m,n)$. Specifically, $\Delta y_k^l(m,n) = \frac{\partial E}{\partial y_k^l(m,n)}$ |
| $\Delta\alpha_k^i, \Delta\beta_k^i$ | Given the cost function, $E$, the individual sensitivities of the bias pair. Specifically, $\Delta\alpha_k^i = \frac{\partial E}{\partial \alpha_k^i}, \Delta\beta_k^i = \frac{\partial E}{\partial \beta_k^i}$ |
| $\nabla_\Psi P_i^{l+1}$ | The derivative of the pooling function, $P_i^{l+1}$ w.r.t a nodal operator, $\Psi$. Since $P_i^{l+1} = \Sigma$, $\nabla_\Psi P_i^{l+1} = 1$ |
| $\nabla_y \Psi$ | The derivative of a nodal operator, $\Psi$, w.r.t the output feature map, $y$. |
| $\nabla_\alpha \vec{y}$ and $\nabla_\beta \vec{y}$ | The derivatives of the shifted output map, $\vec{y}_k^l$, w.r.t the individual bias elements $(\alpha_k^i, \beta_k^i) \in \mathbb{R}[\pm\boldsymbol{\Gamma}]$ |
| $\Delta\alpha_k^i, \Delta\beta_k^i$ | Given the cost function, $E$, the individual sensitivities (derivatives) of the bias elements, $(\alpha_k^i, \beta_k^i) \in \mathbb{R}[\pm\boldsymbol{\Gamma}]$. |

Table 1 presents the formula abbreviations and mathematical symbols used in this article. Back-Propagation (BP) training for Self-ONNs[1] is briefly formulated in Appendix A and further details can be obtained from [10].

## III. SUPER NEURONS WITH NON-LOCALIZED KERNEL OPERATIONS

The starting point of this study is the generative neuron model of Self-ONNs [10]. Like its predecessors, ONNs, and CNNs, each kernel connection of a generative neuron to the previous layer output maps is localized, i.e., for a pixel located at $(m,n)$ in a neuron at the current layer, all kernels are located (centered) at the same location over the previous layer output maps. Figure 2 (bottom-left) illustrates this with $3 \times 3$ kernels where a pixel of the $i^{th}$ neuron in layer $l+1$, $x_i^{l+1}(m,n)$, is computed using the 9 pixels of the previous layer output maps, $y_k^l(m+r, n+t) \; \forall r,t \in [-1,1]$, for $\forall k \in [1, N_l]$ operated with the kernels centered at the same location, $(m,n)$, given that $N_l$ is the number of neurons in the previous layer, $l$. This gives rise to an obvious limitation since such a static kernel is blinded to the neighboring pixels outside of the kernel boundaries, $\forall r,t \in [-1,1]$ which may have the potential to contribute to the input pixel, and hence should not be excluded. This study provides two feasible solutions by proposing two super neuron models with *non-localized* kernel operations as illustrated at the bottom of the figure. We define two additional parameters, $(\alpha_k^i, \beta_k^i)$ as the spatial bias, that is the shift of the kernel from the pixel location, $(m, n)$, towards x- and y-direction. The spatial bias is, therefore, defined for each kernel for both super-neuron models, i.e., in Figure 2, for the first model, the bottom-left illustration shows the shifted kernel locations for the $i^{th}$ neuron input map at layer $l$+1 connected to the $k^{th}$ output neuron at layer $l$, and the bias values are $[\alpha_k^i, \beta_k^i] \in \mathbb{Z}[\pm\Gamma]$ where the maximum range is determined by the hyperparameter, $\Gamma = 4$ pixels. Therefore, all $3 \times 3$ kernels are *randomly* located within a bias range of $[-4,4]$, and thus, all pixels within the region of 11x11 pixels can contribute. In the illustration, different colored kernels are for different connections and their corresponding bias values within the 11x11 region (the outer, red-dashed square) are *randomly* set in advance. For instance, the bias for the 1st connection (black) is, $\alpha_1^i$=4, $\beta_1^i$=3 pixels whereas for the 3rd connection (red), it is $\alpha_3^i$=0, $\beta_3^i$=0, respectively.

---

[1] The optimized PyTorch implementation of Self-ONNs with generative and super neurons is publicly shared in http://selfonn.net/ .

Finally, for the second model, the illustration at the bottom-right shows the shifted kernel locations by the real-valued bias, $(\alpha_k^i, \beta_k^i) \in \mathbb{R}[\pm\Gamma]$. For this super-neuron model, the bias is iteratively optimized during BP training along with other network parameters. At the end of the training, the bias will converge to a (local) optimum point. So, the bottom-right illustration only shows instantaneous localizations of the kernels at a particular BP iteration.

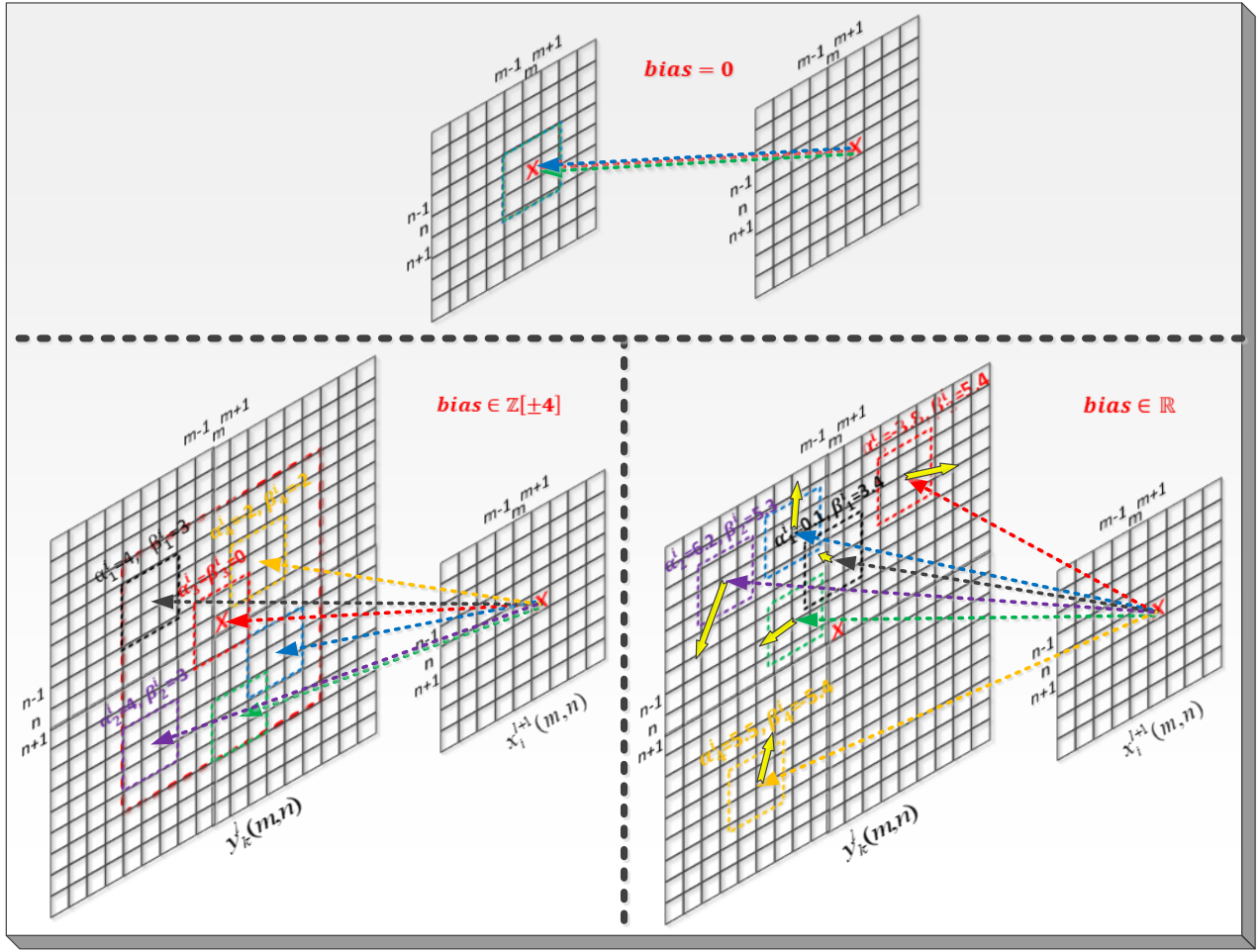

**Figure 2:** Localized (top) *vs.* non-localized kernel operations (bottom) to create the pixel, $x_i^{l+1}(m, n)$, from the output maps of the previous layer neurons. At the (bottom) right, *randomly localized* (uniformly distributed) kernels within a spatial *bias* range of $\Gamma = 4$ are shown. At the (bottom) left, the *BP-optimized* locations of each kernel during a BP epoch with bias gradients, $(\Delta\alpha_k^i, \Delta\beta_k^i)$ (yellow vectors) are illustrated.

To formulate a non-localized kernel for the $i^{th}$ neuron in layer $l$+1, connected to the $k^{th}$ neuron in layer $l$ with a bias in x- and y-directions, $\alpha_k^i$ and $\beta_k^i$, respectively, we will instead shift the output feature map in the previous layer in the opposite direction. For this purpose, let $\mathbf{T}^{(\alpha_k^i, \beta_k^i)}$ be the shift operator over $y_k^l$ by the bias, $[\alpha_k^i, \beta_k^i]$. First, we can perform the (opposite direction) shift to obtain, $y_k^l(m + \alpha_k^i, n + \beta_k^i)$ and then operate with the original Kx × Ky kernel, $w_{ik}^{l+1}$ in a localized manner as expressed in Eq. (3). For generative neurons of Self-ONNs recall that $\mathbf{\Psi}$ is the composite nodal function which is the $Q^{th}$ order Mac-Laurin series. Over the 1D array of kernel elements, $w_{ik}^{l+1}(r, t)$, $\mathbf{\Psi}$ is expressed in Eq. (4) where the DC bias term, $w_{ik}^{l+1}(r, t, 0)$, is omitted. Therefore, each generative neuron has a 3D kernel matrix where the $q^{th}$ coefficient of the kernel element $(r, t)$ is represented by $w_{ik}^{l+1}(r, t, q)$.

In the next sub-sections, we formulate the forward-propagation (FP) for the two super-neuron models each of which performs non-localized kernel operations, the former with *random* bias and the latter with learnable (real-valued) bias through BP. The details of BP training are covered in Appendices B and C.

$$x_i^{l+1} = b_i^{l+1} + \sum_{k=1}^{N_l} \boldsymbol{oper2D}\left(\mathbf{T}^{(\alpha_k^i, \beta_k^i)}(y_k^l), w_{ik}^{l+1}, 'NoZeroPad'\right)$$

$$x_i^{l+1}(m,n)\Big|_{(0,0)}^{(M-1,N-1)} = b_i^{l+1} + \sum_{k=1}^{N_l} \left(P_i^{l+1}\left[\Psi(y_k^l(m+\alpha_k^i, n+\beta_k^i), w_{ik}^{l+1}(0,0)), \ldots, \Psi(y_k^l(m+\alpha_k^i+r, n+\beta_k^i+t), w_{ik}^{l+1}(r,t)), \ldots\right]\right) \quad (3)$$

$$\forall r, t \in [0, Kx-1], [0, Ky-1]$$

$$\Psi\left(y_k^l(m+\alpha_k^i+r, n+\beta_k^i+t), w_{ik}^{l+1}(r,t)\right)$$

$$= w_{ik}^{l+1}(r,t,1)y_k^l(m+\alpha_k^i+r, n+\beta_k^i+t) + w_{ik}^{l+1}(r,t,2)y_k^l(m+\alpha_k^i+r, n+\beta_k^i+t)^2 \quad (4)$$

$$+ \cdots + w_{ik}^{l+1}(r,t,Q)y_k^l(m+\alpha_k^i+r, n+\beta_k^i+t)^Q$$

### A. FP for Non-localized Kernel Operations by the Random Bias

In this super neuron model, the spatial bias consists of an integer pair of shifts in x- and y-directions, $(\alpha_k^i, \beta_k^i) \in \mathbb{Z}[\pm\mathbf{\Gamma}]$, each of which is randomly created within the range limit, $\mathbf{\Gamma}$, for each kernel connection (to each output map in the previous layer) of each super neuron in the network. Therefore, when a Self-ONN is composed of super neurons that are configured with the non-localized kernel operations by *random* spatial bias, each super neuron will have an array of spatial bias pairs that are randomly assigned in advance and used thereafter as the network parameters. During the FP, the native output map(s) of the input layer are acquired from the training data and the 2D shifted output map(s) by the spatial bias, $(\alpha_k^i, \beta_k^i)$ are computed as follows:

$$\vec{y}_k^l = \mathbf{T}^{(\alpha_k^i, \beta_k^i)}(y_k^l) \quad (5)$$

where $l = 0$ for the input layer. Then using Eq. (3) each input map in the next hidden layer, $x_i^{l+1}, \forall i \in [1, N_{l+1}]$, can be computed. Passing the input map through the activation operator, $f(x)$, first and then the pooling (if up- or down-sampling is performed in that layer), the output map, $y_i^{l+1}$, is created. Once again, using Eq. (5), the shifted output map is created, and the FP proceeds to the next layers. To accommodate such shifts, the boundaries of each output map are zero-padded by $\mathbf{\Gamma}$-zeros. In order to speed-up both FP and BP, the $q^{\text{th}}$ power of the shifted outputs, $(\vec{y}_k^l)^q$, can also be computed only once (during FP) and stored to be used repeatedly during BP. On the other hand, except for the output maps of the super neurons in the output layer, there is no need to store the original outputs, $y_k^l$, along with their powers since they are only temporarily needed for non-localized kernel operations.

### B. FP for Non-localized Kernel Operations by the BP-optimized Bias

The BP optimization of each of the pair of bias shifts in x- and y-directions requires that $(\alpha_k^i, \beta_k^i) \in \mathbb{R}$ and thus, the individual gradients (sensitivities), $\Delta\alpha_k^i = \frac{\partial E}{\partial \alpha_k^i}, \Delta\beta_k^i = \frac{\partial E}{\partial \beta_k^i}$, can be computed. In this case, the bias range, $\mathbf{\Gamma}$, can still be defined for practical reasons (e.g., $(\alpha_k^i, \beta_k^i) \in \mathbb{R}[\pm\mathbf{\Gamma}]$) so that the boundaries of each output map can be zero-padded by $\Gamma$-zeros in advance to accommodate the shifts within the allocated memory of the map. As in the random bias approach, during the FP, when an output map, $y_k^l$, is generated by activating the input map, to create each input map in the next layer, $x_i^{l+1}, \forall i \in [1, N_{l+1}]$, using Eq. (3), it will first be shifted by the bias (set earlier by BP) before the 2D (nodal) operation, i.e., $y_k^l \to \mathbf{T}^{(\alpha_k^i, \beta_k^i)}(y_k^l)$. However, since $(\alpha_k^i, \beta_k^i) \in \mathbb{R}$, the shifted map, $\vec{y}_k^l = \mathbf{T}^{(\alpha_k^i, \beta_k^i)}(y_k^l)$, will exist in the fractional grid as illustrated in Figure 3. Since the shifted map location, $\vec{y}_k^l(m+r, n+t)$, is not on the integer grid, but somewhere within the surrounding four grid locations, $(m_\alpha^r, n_\beta^t), \ldots, (m_\alpha^r+1, n_\beta^t+1)$ where $(m_\alpha^r = m + \lfloor\alpha_k^i\rfloor + r, n_\beta^t = n + \lfloor\beta_k^i\rfloor + t)$ and $\forall r, t \in [0, Kx-1], [0, Ky-1]$ are the kernel indices. One can use bilinear (over 4 pixels) or bicubic (over 16 pixels) interpolation to express the shifted map location, $\vec{y}_k^l(m+r, n+t)$. For simplicity and speed, bilinear interpolation is used as expressed in Eq. (6). Hence Eq. (3) can now be modified for FP with the shifted map location, $\vec{y}_k^l(m+r, n+t)$ by the fractional bias as in Eq. (7).

$$\vec{y}_k^l(m+r, n+t) = y_k^l(m_\alpha^r, n_\beta^t)(1-\zeta\alpha_k^i)(1-\zeta\beta_k^i) + y_k^l(m_\alpha^r+1, n_\beta^t+1)\zeta\alpha_k^i\zeta\beta_k^i + y_k^l(m_\alpha^r+1, n_\beta^t)\zeta\alpha_k^i(1-\zeta\beta_k^i) + y_k^l(m_\alpha^r, n_\beta^t+1)(1-\zeta\alpha_k^i)\zeta\beta_k^i \quad (6)$$

$$x_i^{l+1} = b_k^l + \sum_{k=1}^{N_l} \boldsymbol{oper2D}(w_{ki}^l, \vec{y}_k^l, 'NoZeroPad')$$

$$x_i^{l+1}(m,n)\Big|_{(0,0)}^{(M-1,N-1)} = b_k^l + \sum_{k=1}^{N_l} \left(P_i^{l+1}\left[\Psi(\vec{y}_k^l(m,n), w_{ik}^{l+1}(0,0)), \ldots, \Psi(\vec{y}_k^l(m+r, n+t), w_{ik}^{l+1}(r,t)), \ldots\right]\right) \quad (7)$$

where $\zeta\alpha_k^i = \alpha_k^i - \lfloor\alpha_k^i\rfloor$ and $\zeta\beta_k^i = \beta_k^i - \lfloor\beta_k^i\rfloor$.

Note that Eq. (7) is identical to Eq. (1), the one for generative neurons except that the shifted (interpolated) map, $\vec{y}_k^l$, is now used, which is different from the original output map over which a bilinear interpolation is performed. This is in fact equivalent to a low-pass filtering operation over the actual output maps.

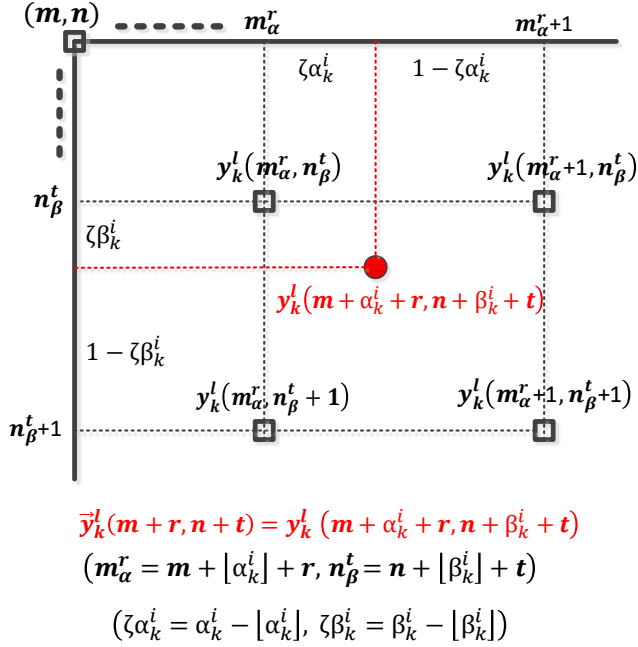

**Figure 3: The composition of the shifted map, $\vec{y}_k^l(m+r, n+t)$ = $y_k^l(m + \alpha_k^i + r, n + \beta_k^i + t)$, by the bias, $(\alpha_k^i, \beta_k^i) \in \mathbb{R}$.**

The BP formulations of the two super neuron models are covered in Appendices B and C, respectively.

## IV. RESULTS

In this section, we have tested Self-ONNs with super neurons against both deep and shallow CNN models over three challenging applications. In the next subsection, we performed real-world image denoising experiments over the SIDD Medium benchmark dataset [63] to perform comparative evaluations against the deep (17-layer) Residual CNN (DnCNN [2]) and (DnONN [1]). For the comparisons using the shallow models, the two super neuron models proposed in this study will then be evaluated against the generative neurons, conventional and deformable [22], [23] convolutional neurons over the following challenging problems: 1) Motion and Spatial Deblurring, and 2) Face Segmentation. Finally, to validate the super neurons' ability to learn the true shift, a "Proof-of-Concept" experimentation using a Self-ONN with only one hidden super-neuron will be presented in Appendix D.

### A. Real-World Denoising

In real-world denoising experiments, we utilize the SIDD Medium training dataset [63] which consists of 320 high-resolution images. We use the same cropping strategy as adopted in [9] to extract 160k training patches. For testing, the SIDD validation dataset is used which consists of 1280 noisy clean image pairs. In all the experiments, the training-to-validation ratio is set to 9:1. For BP training, we use the ADAM optimizer with the maximum learning rate set to $10^{-3}$. All the networks were trained for 100 epochs and the model state which maximized the validation set performance was chosen for evaluation. Model architectures were defined using FastONN [8] library and Pytorch library [64]. All experiments were performed either on an NVIDIA Tesla V100 or an NVIDIA TITAN RTX GPU.

The quantitative results for the real-world denoising problem in terms of average PSNR levels are presented in Table 2 and the visual results on the SIDD Validation dataset are shown in Figure 4. In order to test the hyper-parameter variations in Super-ONN models over the performance, we have trained 5 Super-ONN models:

1) *Super-ONN (Q=3)*: 3-layer Self-ONN with super neurons and *tanh* activation functions.
2) *Super-ONN (Q=2) with ReLU*: 3-layer Self-ONN with super neurons and ReLU activation functions.
3) *Super-ONN (Q=3) LR-IN*: 3-layer Self-ONN with super neurons and ReLU activation functions followed by instance normalization.
4) *Super-ONN (Q=2) Residual LR-IN*: 3-layer Self-ONN with super neurons, ReLU activation functions followed by instance normalization and a residual input-output connection.
5) *Super-ONN (Q=2) Reflection ReLU*: 3-layer Self-ONN with super neurons, ReLU activation functions, and reflection padding in the borders of the images.

**Table 2: Average PSNR levels of real-world denoising for deep models, DnCNN and Dn-SelfONN with 17 layers [2], against the Super-ONNs with 3 hidden layers over the SIDD Medium benchmark dataset.**

| Model | Param. No. | PSNR (dB) |
|---|---|---|
| *Super-ONN (Q=3 with tanh)* | 231K | 37.28 |
| *Super-ONN (Q=2 with ReLU)* | **154K** | 37.32 |
| *Super-ONN (Q=3 LR-IN with tanh)* | 231K | 37.16 |
| *Super-ONN (Q=2 Residual LR-IN)* | **154K** | 37.58 |
| **Super-ONN (Q=2 Reflection ReLU)** | 177K | **37.81** |
| *DnCNN (ReLU)* | 556K | 36.05 |
| *DnONN (Q=3 with tanh)* | 784K | 36.95 |

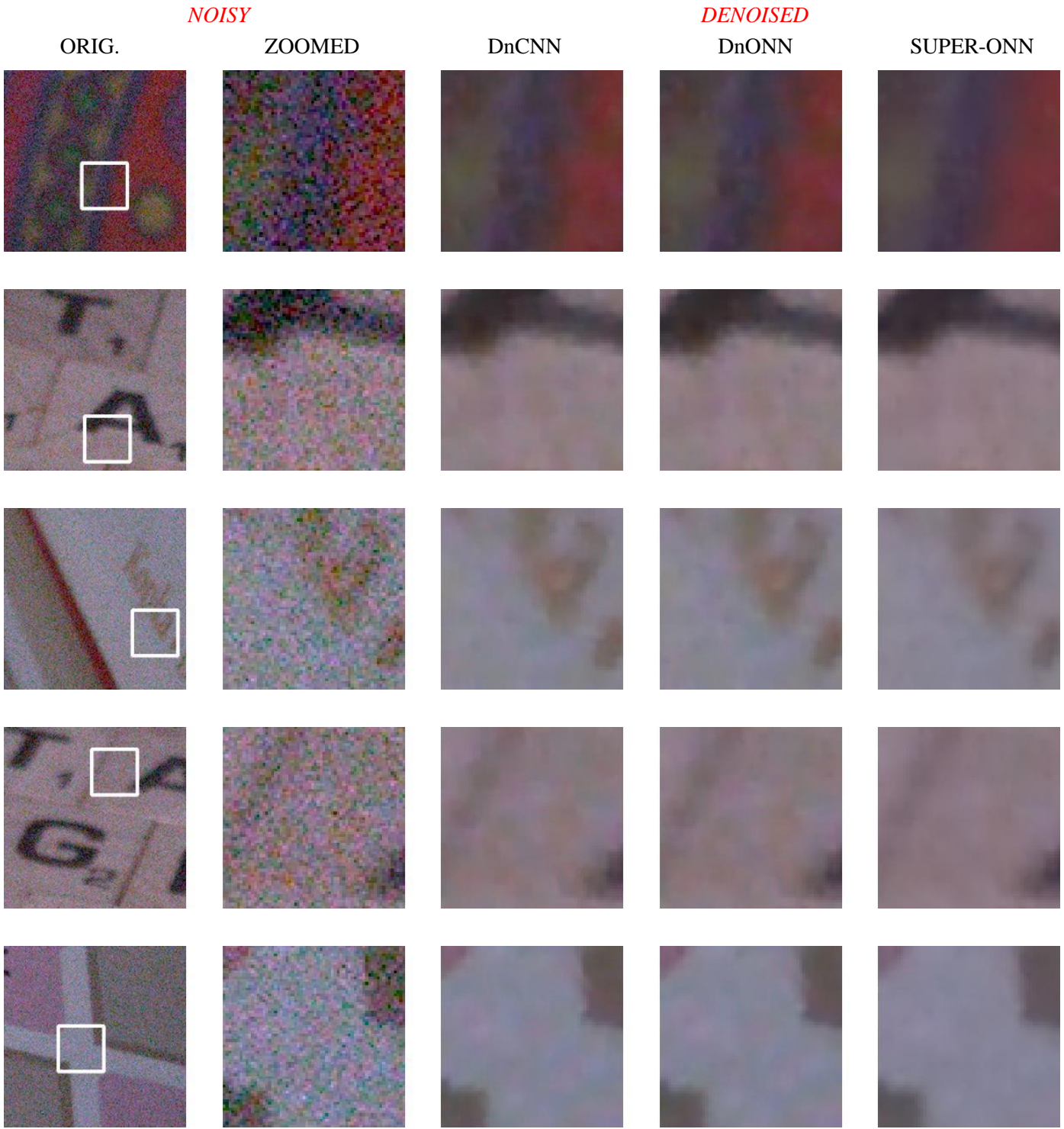

Figure 4: Some sample images with real-world noise (left), their zoomed sections (2nd column), and the corresponding outputs of the DnCNN (3rd column), DnONN (4th column), and Super-ONN (right) from the validation set of SIDD dataset.

The results in Table 2 clearly show that all Super-ONN models significantly outperform both DnCNN and Dn-SelfONN models regardless of their model variations. Especially the performance gap over DnCNN exceeds 1.7dB in PSNR despite the fact that it has more than 5 times more layers and neurons. This demonstrates the superior learning capability of the super neurons over both generative and convolutional neurons.

Qualitatively speaking, the superior denoising performance of Self-ONNs with super-neurons (Super-ONNs) is once again visible in all output images shown in Figure 4. Super-ONNs not only achieve a sharper edge and texture restoration but also recover the smooth regions better than the other networks.

### B. Results with Shallow Models

To test the true learning capability of the super neurons, especially against generative neurons, we further apply the following severe restrictions and harsh conditions:
i) Very Low Resolution: 60x60 pixels,
ii) Compact/Shallow Models with only 2 hidden layers and less than 25 neurons: *Inx12x12xOut*,
iii) Scarce Train Data: only 10% of the dataset is for training and the rest is for testing (10xfold cross-validation)
iv) Limited kernel size (3x3 kernels except for the 2nd problem)

For all problems, restriction (ii) is relaxed for the conventional and deformable CNNs that have a configuration $In \times 48 \times 48 xOut$, hence labeled as "CNN×4" and "DefCNN×4". With 4 times more neurons than Self-ONNs (*Inx12x12xOut*), such an unfair comparative evaluation is intended to show the true learning capability of the super neurons. For Self-ONNs, $Q = 3, 5$, and $7$ at the 1st, 2nd hidden, and output layers, respectively. Moreover, the first hidden layer applies sub-sampling by $ssx = ssy = 2$, and the second one applies up-sampling by $usx = usy = 2$. For each regression problem, we used the Signal-to-Noise Ratio (SNR) evaluation metric, which is defined as the ratio of the signal power to the noise power, i.e., $SNR = 10\log(\sigma_{signal}^2/\sigma_{noise}^2)$. For the *Image Transformation* problem, we performed 10 experiments each with 4 images transformed to another 4. For *Deblurring* and *Denoising*, the benchmark datasets are partitioned into the train (10%) and test (90%) for 10-fold cross-validation. For each fold, all networks are trained using Stochastic Gradient Descent (SGD) with a fixed learning parameter as presented in Table 4. Finally, 5 BP runs are performed and the network model that achieved the minimum loss (MSE) during these runs is used for evaluation (tested over the rest of the dataset).

#### 1) Deblurring

Image deblurring [24]-[36] can broadly be categorized as kernel-based estimation [24]-[30], or an end-to-end system [31]-[36]. Deep CNNs have been used for each category but in this study, we shall evaluate the networks in an "end-to-end" configuration, that does not have to estimate the blurring kernel, rather the blurred image is directly transformed into the restored (deblurred) image. We expect that super neurons with non-localized kernel operations to achieve a superior performance because image deblurring usually requires a large receptive field for enhanced global knowledge [37] while conventional CNNs (and Self-ONNs) can provide local knowledge limited with the size of their filters. We consider two blurring problems:
- *Disc(ρ)* blurring: a circular averaging filter (pillbox) within the square matrix of size, $2 \times \rho + 1$.
- *Motion(Λ, θ)* blurring: the linear motion of a camera where $Λ$ specifies the length of the motion and $θ$ specifies the angle of motion in degrees in a counter-clockwise direction.

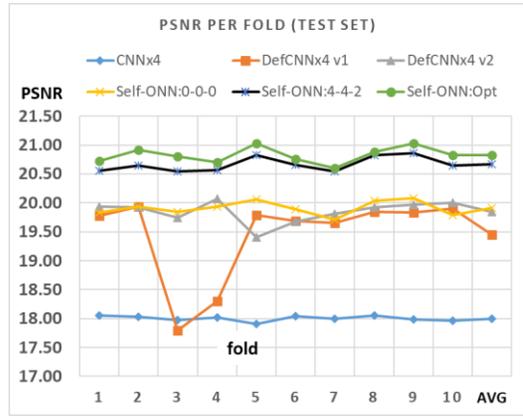
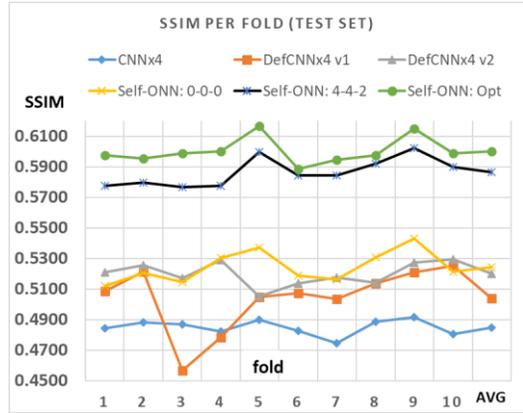
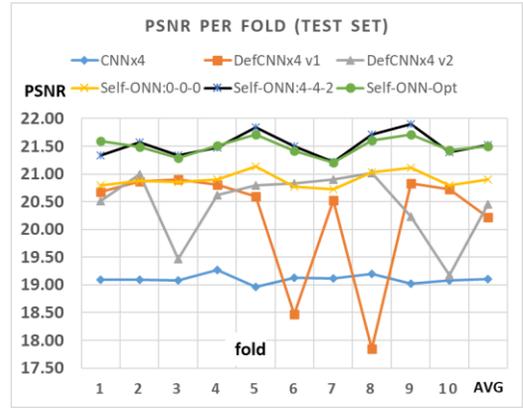
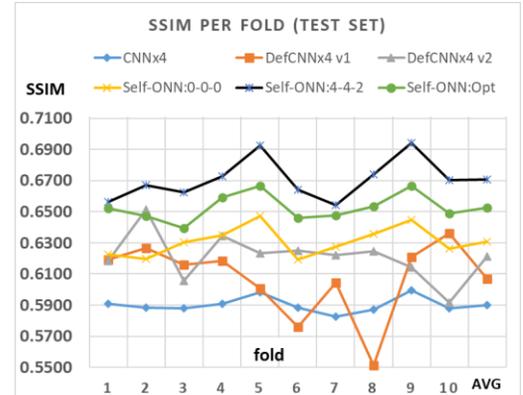

**Figure 5: Best PSNR and SSIM scores for each *Disc-5* (top) and *Motion* (bottom) deblurring fold achieved by the corresponding Self-ONNs (with no, random and BP-optimized spatial biases) and the three CNN×4 configurations over the test set.**

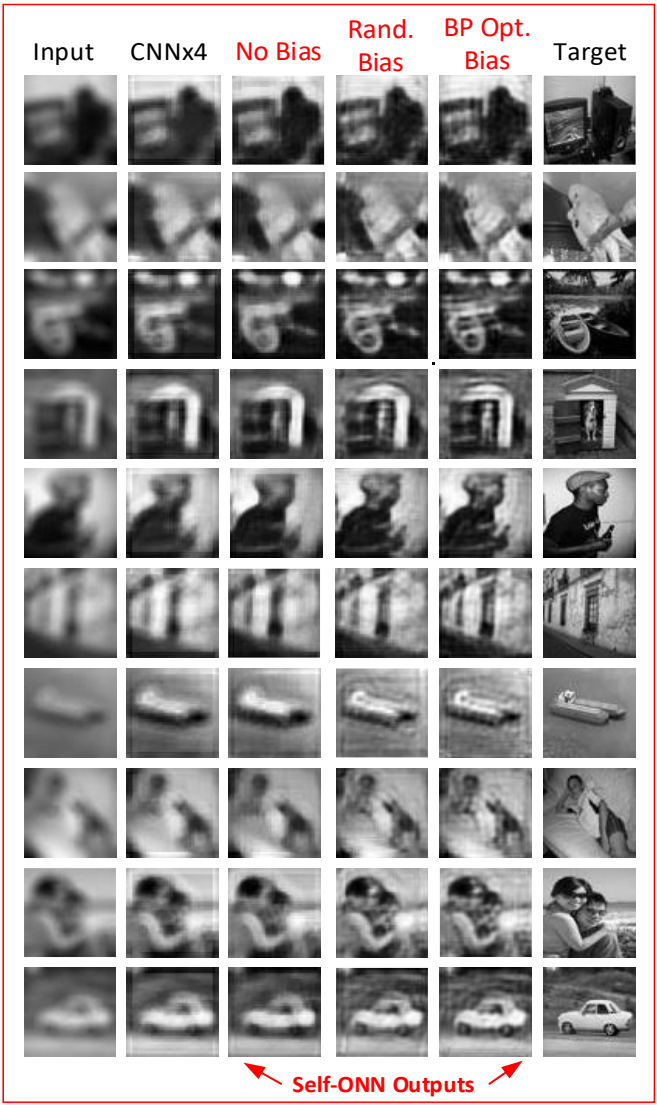

**Figure 6:** Some typical original (target) and *Disc-5* blurred (input) images and the corresponding outputs of the CNN×4 and the three Self-ONNs (with no, random and BP-optimized spatial bias) from the test partition.

In both problems, we aim to evaluate the learning capability of the super neurons in Self-ONNs against the generative, conventional, and especially deformable convolutional neuron models under harsh conditions and for this reason, along with the aforementioned restrictions, we apply a severe blurring with the following parameter settings: $\rho = 5, \Lambda = 11\ and\ \Theta = \pi/4$. $Disc(\rho)$ basically applies an averaging of $11 \times 11$ pixels and $Motion(\Lambda, \Theta)$ approximates a linear motion of 11 pixels diagonally. Both blurring artifacts can sometimes cause such severe image degradation that makes it difficult or even infeasible to comprehend the content of the image (e.g., see Figure 6). Finally, for the random bias ranges for super neurons where $(\alpha_k^i, \beta_k^i) \in \mathbb{Z}[\pm\mathbf{\Gamma}]$, are set as $\mathbf{\Gamma} = \{4,4,2\}$ for the 1st, 2nd, and output layers, respectively. Thus, with this setting, for instance, the 1st layer super neurons will have the improved size of the receptive fields as 11x11 pixels, which is significantly larger than the original kernel size of 3x3 pixels.

Figure 5 shows PSNR and SSIM plots of the best *Disc-5* (top) and motion (bottom) deblurring results per fold over the test partitions. The Self-ONNs with generative neurons having no bias are with the label '0-0-0'), and with super neurons having random bias within $\mathbf{\Gamma} = \{4,4,2\}$ are with the label '4-4-2', and with BP-optimized bias are with the label 'Opt', respectively. The conventional and the two versions of the deformable CNNs have the labels 'CNN×4', 'DefCNN×4 v1', and 'DefCNN×4 v2', respectively. The average PSNR and SSIM scores are presented at the end of each corresponding plot.

In both problems, Self-ONNs achieve significantly higher PSNR (around 1dB) and SSIM (> 4%) levels as compared to the three CNN models with four times more neurons. Though both deformable CNNs achieve slightly higher PSNR and SSIM scores than the conventional CNNs in the majority of the folds, they fail to achieve a higher average performance due to the lowest scores obtained in two folds, indicating a robustness issue. Finally, in both problems, the Self-ONNs with super neurons achieve more than 0.6dB higher PSNR score on average compared to Self-ONNs with generative neurons.

For a visual evaluation, Figure 6 and Figure 7 show a set of *Disc-5* and *motion*-blurred (input) images, the target image, and the corresponding outputs of CNN×4, Self-ONNs (with no, random, and BP-optimized spatial biases) from the test partition. We skipped the outputs from both deformable CNNs since they have a very similar or occasionally worse visual quality than the conventional CNN×4 model. The superior deblurring performance of Self-ONNs with super neurons is visible in all outputs.

*2) Face Segmentation*

Deep CNNs have often been used in face and object segmentation tasks [52]-[61]. In this study, we use the benchmark FDDB face detection dataset [44], which contains 2000 images with one or many human faces in each image. As per the aforementioned restriction, all images are down-sampled to 60x60 pixels and in this very low resolution, pixel-accurate face segmentation becomes an even more challenging task.

Finally, for the random bias ranges for super neurons where $(\alpha_k^i, \beta_k^i) \in \mathbb{Z}[\pm\mathbf{\Gamma}]$, are set as $\Gamma = \{4,4,2\}$ for the 1st, 2nd and output layers, respectively. Thus, with this setting, for instance, the 1st layer super neurons will have the improved size of the receptive fields as 11x11 pixels, which is significantly larger than the original kernel size of 3x3 pixels. Figure 8 shows F1 plots of the best (in training) *Face Segmentation* results per fold over the test set. The average test F1 scores achieved by the three Self-ONNs are 80.95% (no bias), 83.83% (random bias), 84.22% (BP-optimized bias), respectively whilst the CNN×4 has the F1-score of 75.94%. In both train and test partitions and all folds Self-ONNs achieve significantly higher performance as compared to CNNs. This is despite the fact that it has four times less neurons. In particular, the average performance gap between CNNs and Self-ONNs with super neurons is widened around 8% and 5.6% in train and test partitions, respectively. Finally, the Self-ONNs with super neurons can achieve higher than 3% (train) and 4% (test) on the average than the corresponding performance of the Self-ONNs with generative neurons.

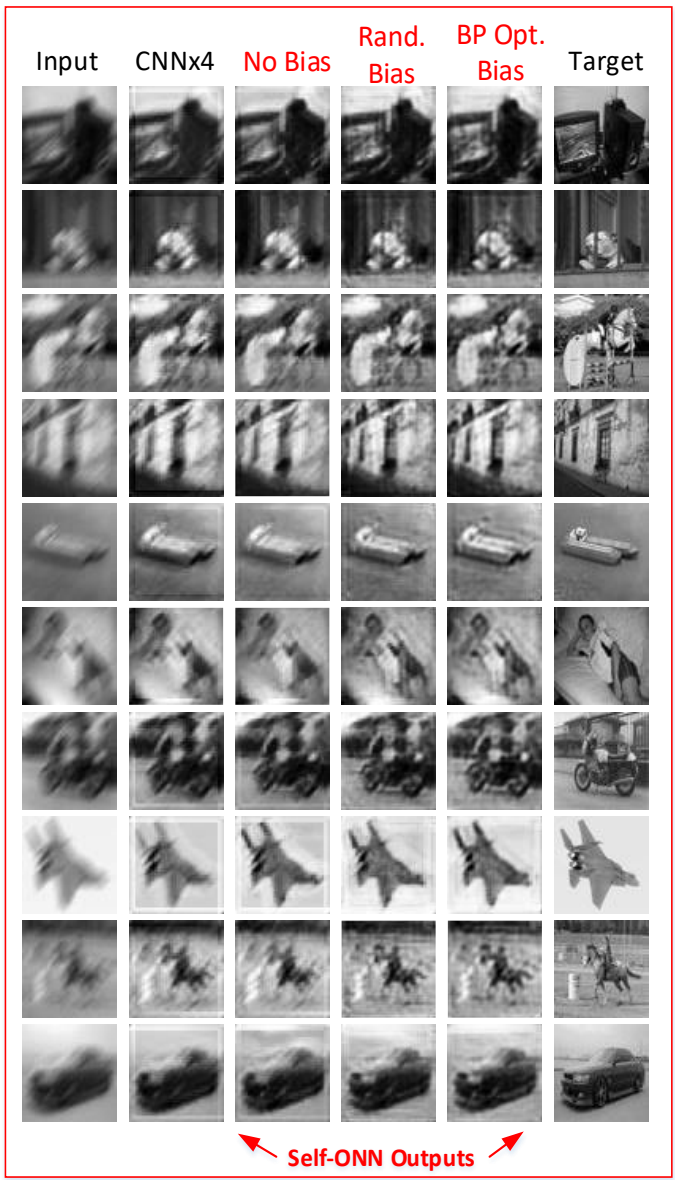

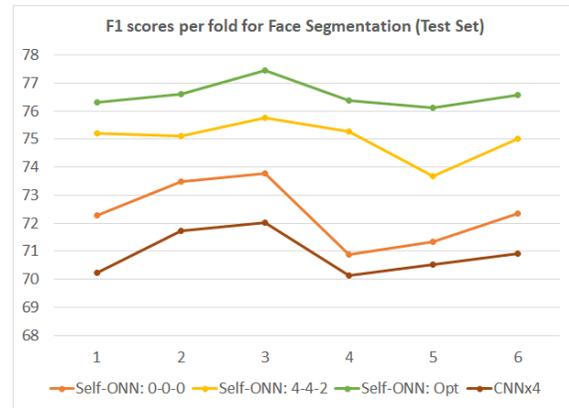

**Figure 8:** Best F1 scores for each *face segmentation* fold achieved by the corresponding Self-ONN (with no, random and BP-optimized spatial biases) and CNNx4.

**Figure 7:** Some typical original (target) and *Motion* blurred (input) images and the corresponding outputs of the CNN×4 and the three Self-ONNs (with no, random and BP-optimized spatial bias) from the test partition.

For a visual evaluation, Figure 9 shows some typical original input images (first column), their (target) ground-truth face maps (last column), and the corresponding outputs of the CNN×4 and the three Self-ONNs (with no, random and BP-optimized spatial bias) from the test partition. Obviously, the best face segmentation results belong to the Self-ONNs with super neurons while CNNs suffer from severe false-positive regions. The super neurons with BP-optimized spatial bias yield the overall best results with minimal false positives

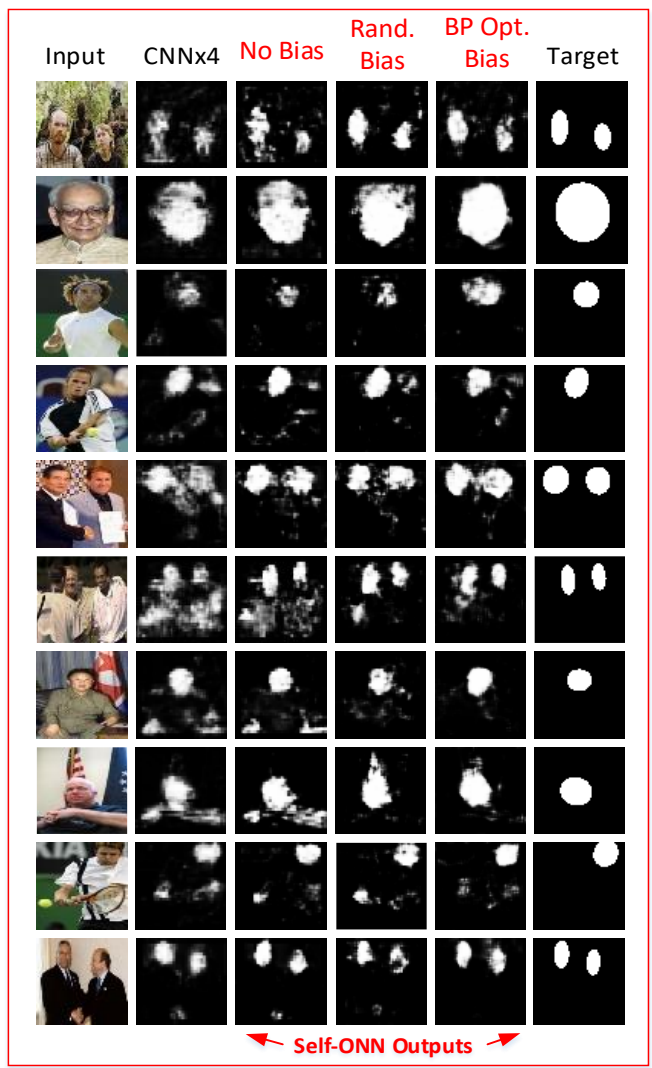

**Figure 9:** Some typical original input images (first column) their (target) ground-truth face maps (last column) and the corresponding outputs of the CNN×4 and the three Self-ONNs (with no, random and BP-optimized spatial bias) from the test partition.

## C. Computational Complexity Analysis

In this section, the computational complexity of the proposed Self-ONNs with super neurons is analyzed with respect to the parameter-equivalent Self-ONNs with generative neurons and the three CNN models with 4 times more neurons. As assumed in this study, when the pool operator is the "summation", $P_i^l = \Sigma$, in an FP of a Self-ONN with super neurons, Eq. (3) can be expressed as follows:

$$x_k^l = b_k^l + \sum_{i=1}^{N_{l-1}} \boldsymbol{oper2D}(\vec{y}_i^{l-1}, w_{ki}^l, 'NoZeroPad')$$

$$x_k^l(m,n)\Big|_{(0,0)}^{(M-1,N-1)} = b_k^l + \sum_{i=1}^{N_{l-1}}\left(\sum_{r=0}^{K_x-1}\sum_{t=0}^{K_y-1}\boldsymbol{\Psi}(\vec{y}_i^{l-1}(m+r,n+t), \boldsymbol{w_{ki}^l(r,t)})\right) \quad (8)$$

where $\boldsymbol{\Psi}$ is the (Taylor series) nodal operator function and $w_{ki}^l(\mathbf{r,t})$ is a $Q$-dimensional array for the kernel element $(\mathbf{r,t})$. Putting the $q^{th}$ order 2D kernel, $w_{ki}^l\langle q \rangle$ ($q=1..Q$), which is composed of the *kernel elements*, $w_{ik}^{l+1}(r,t,q)$, then Eq. (8) can be simplified as,

$$x_k^l = b_k^l + \sum_{q=1}^{Q}\left\{\sum_{i=1}^{N_{l-1}} conv2D((\vec{y}_i^{l-1})^q, w_{ki}^l\langle q \rangle, 'NoZeroPad')\right\} \quad (9)$$

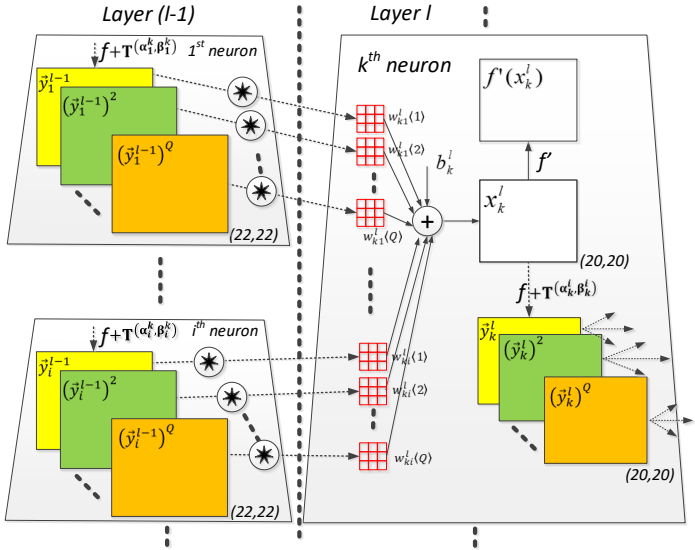

**Figure 10:** The illustration of a Self-ONN equivalent to Figure 1 (right) when the pool operator is "sum", $P_i^l = \Sigma$, and the activation function is *tanh*.

Such a 2D convolutional representation of a generative neuron's input map formation is illustrated in Figure 10. It is straightforward to see that this indeed resembles a multi-output and multi-kernel convolutional neuron. When the shifted powers of the output maps, $(\vec{y}_1^{l-1})^q$, for $q=1,..., Q$, are computed for all hidden neurons in the network, Eq. (9) simply turns out to be $(Q \times N_{l-1})$ independent 2D convolutions. Like in conventional CNNs, this can be implemented in a parallel manner, and hence, it will roughly take a similar inference time. We can thus conclude that, in a parallelized implementation, a Self-ONN and a CNN with the same configuration have similar computational complexity. For both super neuron models, the number of parameters, $PARs$, in the Self-ONNs can be expressed as,

$$PARs = \sum_{1}^{L} PARs(l)$$

$$= \sum_{1}^{L}\Big(\big((N_{l-1} \times (K_x^l \times K_y^l \times Q_l + 2)) + 1\big) * N_l\Big) \quad (10)$$

**Table 3: Comparison of the total number of multiply-accumulate operations for the networks used in this study.**

| Network | Layer 1 Size | Layer 2 Size | *No. of* $PARs$ (K) | Memory Overhead (Kb) |
|---|---|---|---|---|
| SelfONN (no *bias*) | 12 | 12 | 7.585 | 0 |
| SelfONN (*rand. bias*) | 12 | 12 | 9.265 | 0 |
| SelfONN (*BP-opt. bias*) | 12 | 12 | 9.265 | 0 |
| CNN×4 | 48 | 48 | 21.697 | 0 |
| DefCNN×4 v1 | 48 | 48 | 37.368 | 583.2 |
| Def CNN×4 v2 | 48 | 48 | 45.252 | 874.9 |

For each network model, Table 3 presents the number of network parameters, $PARs$ and the memory overhead, which is the additional memory needed during the FP besides the network parameters and I/O buffers for feature maps. Besides having 4 times more neurons, it is apparent from the table that all CNN×4 models have around 2.3 to 6 times more parameters and 2.2 to 5 times higher computational complexity than the Self-ONNs with super neurons that are configured with the non-localized kernel operations by random spatial bias. Whilst having a similar computational complexity, the only overhead cost for super neurons over the generative neurons is about 1.22 times more parameters due to the spatial bias elements. This is true for both models, randomized and BP-optimized bias; however, super neurons with BP-optimized bias have around 1.1 times higher computational complexity than the models with no bias (generative neurons) and random bias. This is due to the bilinear interpolation performed to compute the shifted output maps.

Particularly, for deformable CNN×4 models, v1 and v2, the memory overhead, $MEM^+$, can be expressed as follows:

$$MEM^+ = \sum_{1}^{L} MEMs(l)$$

$$= \sum_{1}^{L}\Big(K\big(Bs \times Gs \times K_x^l \times K_y^l \times W_x^l \times W_y^l\big)\Big) \quad (11)$$

where $K$ is constant ($K = 2$ for v1 and $K = 3$ for v2), $Bs$ is the batch size, $Gs$ is the group size, $W_x^l$ and $W_y^l$ are the width and height of the input feature map of the layer, $l$. The memory overhead can, therefore, be infeasibly large, especially for deep networks with practical settings. As an example, for a single layer with 256×256 pixel feature maps, 3×3 kernels, and $Gs = Bs = 8$, v1 and v2 versions of deformable CNNs will require around 302Mb and 452Mb extra memory, respectively only for a single layer.

## V. CONCLUSIONS

The ancient neuron model from the 1950s [38] has been used by the MLPs ever since, and later on shared by its popular derivative, the conventional CNNs. As a linear model, it can only perform linear transformations with "localized" kernels making CNNs entirely homogenous with a static neuron model in terms of transformation and localization. This study is inspired by the well-known proverb, "doing the *right* thing at the *right* place and the *right* time". The Self-ONNs with the generative neuron model can do the "right thing" by customizing each nodal operator on the fly. So, the generative neurons can create the best possible operator for the kernel of each connection during BP training. However, generative neurons can neither locate the "*right*" place" for their kernels nor enhance their "receptive field" bounded by the kernel size. To overcome this, the proposed super neurons can be jointly optimized to do the *right* transformation at the *right* (kernel) location of the *right* connection to maximize the learning performance. This study has proposed two models for super neurons: *randomized* and *BP-optimized* kernel localizations of each connection. Both models improve the size of the receptive fields but only the latter one can seek the *right* (kernel) location of each connection. However, we observe that the underlying problem may not require the "right" location and in this case, both approaches are expected to perform either equally well or the former approach can work even slightly better than the latter because it can optimize each nodal operator of each kernel during the entire BP run without altering the location. Whereas the latter approach jointly optimizes both the nodal operator and the location (spatial bias) during the BP run, this is a significantly harder task because the optimization of the nodal operator cannot be finalized while the kernel keeps moving in each BP iteration. In other words, the optimal nodal operator will obviously be different for different kernel locations, and until the location (the spatial bias) is converged the nodal operator optimization cannot be finalized.

Both models of super neurons are evaluated against the conventional and deformable convolutional neurons of CNNs and generative neurons of Self-ONNs. First, shallow Self-ONNs with super-neurons (Super-ONNs in short) have been tested against deep models: DnCNN and DnONN in a Real-World denoising problem. Despite being a significantly shallow model with few neurons, Super-ONNs outperformed both deep models. Then, to reveal the true learning capabilities of super neurons, we purposefully selected challenging learning tasks and applied harsh learning conditions and restrictions such as scarce train data, shallow configurations with few neurons, and minimal kernel size. Despite 4 times more learning units (neurons) being used for all CNN models for comparative evaluations, the results clearly show that Self-ONNs with super neurons can achieve a superior learning and generalization capability thanks to the improved receptive field size they can provide. The computational complexity analysis reveals that an elegant computational efficiency is also achieved in terms of network parameters and memory overhead. In most problems, a notable performance gap is observed over the conventional Self-ONNs with generative neurons without any significant computational burden.

We can foresee that further performance boost can be expected for the Self-ONNs with super-neurons with the following improvements:

- instead of fixing to some naïve values for the two hyper-parameters, Q and **Γ,** we are aiming to optimize each parameter per layer,
- adapting a better optimization scheme for training, e.g., SGD with momentum [39], AdaGrad [40], RMSProp [41], Adam [42] and its variants [43], all of which should be adapted for Super-ONNs for proper functioning,
- and implementing other kernel operations such as scaling and rotation.

These will be the topics for our future research. The optimized PyTorch implementations of Self-ONNs and Super-ONNs are publicly shared in [62].

# APPENDICES

## A. Training by Back Propagation for Self-ONNs with Generative Neurons

For Self-ONNs, the contributions of each pixel in the $M \times N$ output map, $y_k^l(m,n)$ on the next layer input map, $x_i^{l+1}(m,n)$, can now be expressed as in Eq. (12). Using the chain rule, the delta error of the output pixel, $y_k^l(m,n)$, can therefore, be expressed as in Eq. (13) in the generic form of pool, $P_i^{l+1}$, and composite nodal operator function, $\Psi$, of each operational neuron $i \in [1,..,N_{l+1}]$ in the next layer. In Eq. (13), note that the first term, $\frac{\partial x_i^{l+1}(m-r,n-t)}{\partial P_i^{l+1}[..,\Psi(y_k^l(m,n),w_{ik}^{l+1}(r,t)),..]} = 1$.

Let $\nabla_\Psi P_i^{l+1}(m,n,r,t) = \frac{\partial P_i^{l+1}[..,\Psi(y_k^l(m,n),w_{ik}^{l+1}(r,t)),..]}{\partial \Psi(y_k^l(m,n),w_{ik}^{l+1}(r,t))}$ and $\nabla_y \Psi(m,n,r,t) = \frac{\partial \Psi(y_k^l(m,n),w_{ik}^{l+1}(r,t))}{\partial y_k^l(m,n)}$. Then, Eq. (13) simplifies to Eq. (14). Note further that $\Delta y_k^l$, $\nabla_{\Psi_{ki}} P_i^{l+1}$ and $\nabla_y \Psi$ have the same size, $M \times N$ while the next layer delta error, $\Delta_i^{l+1}$, has the size, $(M - K_x + 1) \times (N - K_y + 1)$, respectively. Therefore, to enable this variable 2D convolution in this equation, the delta error, $\Delta_i^{l+1}$, is padded by zeros at all four boundaries ($K_x - 1$ zeros on left and right, $K_y - 1$ zeros on the bottom and top). Thus, $\nabla_y \Psi(m,n,r,t)$ can simply be expressed as in Eq. (15).

$$\begin{aligned}
x_i^{l+1}(m-1,n-1) &= \ldots + P_i^{l+1}\big[\Psi(y_k^l(m-1,n-1),w_{ik}^{l+1}(0,0)),\ldots,\Psi(y_k^l(m,n),w_{ik}^{l+1}(1,1))\big] + \ldots \\
x_i^{l+1}(m-1,n) &= \ldots + P_i^{l+1}\big[\Psi(y_k^l(m-1,n),w_{ik}^{l+1}(0,0)),\ldots,\Psi(y_k^l(m,n),w_{ik}^{l+1}(1,0)),\ldots\big] + \ldots \\
\boldsymbol{x_i^{l+1}(m,n)} &= \ldots + P_i^{l+1}\big[\Psi(y_k^l(m,n),w_{ik}^{l+1}(0,0)),\ldots,\Psi(y_k^l(m+r,n+t),w_{ik}^{l+1}(\mathbf{r,t}),)\ldots\big] + \ldots \\
&\ldots\ldots \\
\therefore x_i^{l+1}(m-r,n-t)\Big|_{(1,1)}^{(M-1,N-1)} &= b_i^{l+1} + \sum_{k=1}^{N_1} P_i^{l+1}\big[\ldots,\Psi(y_k^l(m,n),w_{ik}^{l+1}(r,t)),\ldots\big]
\end{aligned} \tag{12}$$

$$\therefore \frac{\partial E}{\partial y_k^l}(m,n)\Big|_{(0,0)}^{(M-1,N-1)} = \Delta y_k^l(m,n) = \sum_{i=1}^{N_{l+1}} \left( \sum_{r=0}^{K_x-1} \sum_{t=0}^{K_y-1} \frac{\partial E}{\partial x_i^{l+1}(m-r,n-t)} \times \frac{\partial x_i^{l+1}(m-r,n-t)}{\partial P_i^{l+1}[..,\Psi(y_k^l(m,n),w_{ik}^{l+1}(r,t)),..]} \times \frac{\partial P_i^{l+1}[..,\Psi(y_k^l(m,n),w_{ik}^{l+1}(r,t)),..]}{\partial \Psi(y_k^l(m,n),w_{ik}^{l+1}(r,t))} \times \frac{\partial \Psi(y_k^l(m,n),w_{ik}^{l+1}(r,t))}{\partial y_k^l(m,n)} \right) \tag{13}$$

$$\Delta y_k^l(m,n)\Big|_{(0,0)}^{(M-1,N-1)} = \sum_{i=1}^{N_{l+1}} \left( \sum_{r=0}^{K_x-1} \sum_{t=0}^{K_y-1} \Delta_i^{l+1}(m-r,n-t) \times \nabla_\Psi P_i^{l+1}(m,n,r,t) \times \nabla_y \Psi(m,n,r,t) \right)$$

Let $\nabla_y P_i^{l+1}(m,n,r,t) = \nabla_\Psi P_i^{l+1}(m,n,r,t) \times \nabla_y \Psi(m,n,r,t)$, then

$$\Delta y_k^l = \sum_{i=1}^{N_{l+1}} Conv2Dvar\{\Delta_i^{l+1}, \nabla_y P_i^{l+1}(m,n,r,t)\} \tag{14}$$

$$\nabla_y \Psi(m,n,r,t) = w_{ik}^{l+1}(r,t,1) + 2w_{ik}^{l+1}(r,t,2)y_k^l(m,n) + \cdots + Q w_{ik}^{l+1}(r,t,Q) y_k^l(m,n)^{Q-1} \tag{15}$$

Once the $\Delta y_k^l$ is computed, using the chain-rule, one can express,

$$\Delta_k^l = \frac{\partial E}{\partial x_k^l} = \frac{\partial E}{\partial y_k^l}\frac{\partial y_k^l}{\partial x_k^l} = \frac{\partial E}{\partial y_k^l} f'(x_k^l) = \Delta y_k^l f'(x_k^l) \tag{16}$$

When there is a down-sampling by factors, *ssx* and *ssy*, then the back-propagated delta-error should be first up-sampled to compute the delta-error of the neuron. Let zero order up-sampled map be: $uy_k^l = \underset{ssx,ssy}{\text{up}}(y_k^l)$. Then Eq. (16) can be modified, as follows:

$$\begin{aligned}
\Delta_k^l &= \frac{\partial E}{\partial x_k^l} = \frac{\partial E}{\partial y_k^l}\frac{\partial y_k^l}{\partial x_k^l} = \frac{\partial E}{\partial y_k^l}\frac{\partial y_k^l}{\partial uy_k^l}\frac{\partial uy_k^l}{\partial x_k^l} \\
&= \underset{ssx,ssy}{\text{up}}(\Delta y_k^l)\beta f'(x_k^l)
\end{aligned} \tag{17}$$

where $\beta = \frac{1}{ssx.ssy}$ since each pixel of $y_k^l$ is now obtained by averaging (*ssx.ssy*) number of pixels of the intermediate output, $uy_k^l$. Finally, when there is a up-sampling by factors, *usx* and *usy*, then let the average-pooled map be: $dy_k^l = \underset{usx,usy}{\text{down}}(y_k^l)$. Then Eq. (17) can be updated as follows:

$$\Delta_k^l = \frac{\partial E}{\partial x_k^l} = \frac{\partial E}{\partial y_k^l}\frac{\partial y_k^l}{\partial x_k^l} = \frac{\partial E}{\partial y_k^l}\frac{\partial y_k^l}{\partial dy_k^l}\frac{\partial dy_k^l}{\partial x_k^l}$$
$$= \underset{usx,usy}{\text{down}}(\Delta y_k^l)\beta^{-1}f'(x_k^l) \tag{18}$$

As for the computation of the kernel and bias sensitivities, recall the expression between an individual kernel weight array, $w_{ik}^{l+1}(\mathbf{r},\mathbf{t})$, and the input map of the next layer, $x_i^{l+1}(m,n)$:

$$x_i^{l+1}(m,n)\Big|_{(1,1)}^{(M-1,N-1)} = b_i^{l+1} + \sum_{i=1}^{N_{l-1}} P_i^{l+1}\begin{bmatrix}\Psi\left(y_k^l(m,n), w_{ik}^{l+1}(\mathbf{0},\mathbf{0})\right),\ldots,\\ \Psi(y_k^l(m+r,n+t), w_{ik}^{l+1}(\mathbf{r},\mathbf{t}))\ldots)\end{bmatrix} \tag{19}$$

where the $q^{\text{th}}$ element of the array, $w_{ik}^{l+1}(\mathbf{r},\mathbf{t})$, contributes to all the pixels of $x_i^{l+1}(m,n)$. By using the chain rule of partial derivatives, one can express the weight sensitivities, $\frac{\partial E}{\partial w_{ik}^{l+1}}$, in Eq. (20). A close look to Eq. (20) reveals that, $\frac{\partial \Psi\left(y_k^l(m+r,n+t),w_{ik}^{l+1}(r,t)\right)}{\partial w_{ik}^{l+1}(r,t,q)} = y_k^l(m+r,n+t)^q$, which then simplifies to Eq. (21) Note that in this equation, the first term, $\Delta_1^{l+1}(m,n)$, is independent from the kernel indices, $r$ and $t$. It will be element-wise multiplied by the other two latter terms, each with the same dimension $(M-Kx+1)x(N-Ky+1)$, and created by derivative functions of nodal and pool operators applied over the shifted pixels of $y_k^l(m+r,n+t)$ and the corresponding weight value, $w_{ik}^{l+1}(\mathbf{r},\mathbf{t})$.

$$\frac{\partial E}{\partial w_{ik}^{l+1}}(r,t,q)\Big|_{(0,0,1)}^{(Kx-1,Ky-1,Q)} = \sum_{m=0}^{M-Kx+1}\sum_{n=0}^{N-Ky+1} \frac{\partial E}{\partial x_1^{l+1}(m,n)} \times \frac{\partial x_1^{l+1}(m,n)}{\partial P_i^{l+1}\left[\Psi\left(y_k^l(m,n),w_{ik}^{l+1}(\mathbf{0},\mathbf{0})\right),\ldots,\Psi(y_k^l(m+r,n+t),w_{ik}^{l+1}(\mathbf{r},\mathbf{t}))\ldots)\right]} \times \frac{\partial P_i^{l+1}\left[\Psi\left(y_k^l(m,n),w_{ik}^{l+1}(\mathbf{0},\mathbf{0})\right),\ldots,\Psi(y_k^l(m+r,n+t),w_{ik}^{l+1}(\mathbf{r},\mathbf{t}))\ldots)\right]}{\partial \Psi\left(y_k^l(m+r,n+t),w_{ik}^{l+1}(\mathbf{r},\mathbf{t})\right)} \times \frac{\partial \Psi\left(y_k^l(m+r,n+t),w_{ik}^{l+1}(\mathbf{r},\mathbf{t})\right)}{\partial w_{ik}^{l+1}(r,t,q)} \tag{20}$$

where $\frac{\partial x_1^{l+1}(m,n)}{\partial P_i^{l+1}\left[\Psi(y_k^l(m,n),w_{ik}^{l+1}(\mathbf{0},\mathbf{0})),\ldots,\Psi(y_k^l(m+r,n+t),w_{ik}^{l+1}(\mathbf{r},\mathbf{t}))\ldots)\right]} = 1$ and $\frac{\partial \Psi\left(y_k^l(m+r,n+t),w_{ik}^{l+1}(r,t)\right)}{\partial w_{ik}^{l+1}(r,t,q)} = y_k^l(m+r,n+t)^q$

$$\therefore \frac{\partial E}{\partial w_{ik}^{l+1}}(r,t,q)\Big|_{(0,0,1)}^{(Kx-1,Ky-1,Q)} = \sum_{m=0}^{M-Kx}\sum_{n=0}^{N-Ky}\Delta_1^{l+1}(m,n) \times \nabla_\Psi P_i^{l+1}(m+r,n+t,r,t) \times y_k^l(m+r,n+t)^q \tag{21}$$

If $P_i^{l+1} = \Sigma$, then

$$\frac{\partial E}{\partial w_{ik}^{l+1}}(r,t,q)\Big|_{(0,0,1)}^{(Kx-1,Ky-1,Q)} = \sum_{m=0}^{M-Kx}\sum_{n=0}^{N-Ky}\Delta_1^{l+1}(m,n) \times y_k^l(m+r,n+t)^q \tag{22}$$

$$\therefore \frac{\partial E}{\partial w_{ik}^{l+1}}\langle q\rangle = conv2D\left(\Delta_i^{l+1}, (y_k^l)^q, 'NoZeroPad'\right)$$

$$\frac{\partial E}{\partial b_k^l} = \sum_m\sum_n \frac{\partial E}{\partial x_k^l(m,n)}\frac{\partial x_k^l(m,n)}{\partial b_k^l} = \sum_m\sum_n \Delta_k^l(m,n) \tag{23}$$

In Eq. (21) there is no need to register a 4D matrix for $\nabla_w\Psi = y_k^l(m+r,n+t)^q$ since it can directly be computed from the outputs of the neurons. Moreover, when the pool operator is the sum, then $\nabla_\Psi P_i^{l+1}(m,n,r,t) = 1$ and Eq. (21) will simplify to Eq. (22) where $\frac{\partial E}{\partial w_{ik}^{l+1}}\langle q\rangle$ is the $q^{\text{th}}$ 2D sensitivity kernel, which contains the updates (SGD sensitivities) for the weights of the $q^{\text{th}}$ order outputs in Maclaurin polynomial. Finally, the bias sensitivity expressed in Eq. (23) is the same for ONNs and CNNs since the bias is the common additive term for all.

Let $w_{ik}^{l+1}\langle q\rangle$ be the $q^{\text{th}}$ 2D sub-kernel where $q=1..Q$ and composed of *kernel elements*, $w_{ik}^{l+1}(r,t,q)$. During each BP iteration, $t$, the kernel parameters (weights), $w_{ik}^{l+1}\langle q\rangle(t)$, and biases, $b_i^l(t)$, of each neuron in the Self-ONN are updated until a stopping criterion is met. Let, $\varepsilon(t)$, be the learning factor at iteration, $t$. One can express the update for the weight kernel and bias at each neuron, $i$, at layer, $l$ as follows:

$$w_{ik}^{l+1}\langle q\rangle(t+1) = w_{ik}^{l+1}\langle q\rangle(t) - \varepsilon(t)\frac{\partial E}{\partial w_{ik}^{l+1}}\langle q\rangle$$

$$b_i^l(t+1) = b_i^l(t) - \varepsilon(t)\frac{\partial E}{\partial b_i^l} \tag{24}$$

As a result, the pseudo-code for BP is presented in Alg. 1.

**Algorithm 1: BP training for Self-ONNs with generative neurons**

| |
|---|
| **Input**: Self-*ONN*, *Stopping Criteria* (*maxIter*, *minMSE*) |
| **Output**: Self-*ONN\** = *BP(Self-ONN, maxIter, minMSE)* |
| 1) **Initialize** network parameters randomly (i.e., ~U(-a, a)) |
| 2) **UNTIL** a stopping criterion is reached, **ITERATE**: |
|    a. **For** each mini-batch in the train dataset, **DO**: |
|      i. **FP**: Forward propagate from the input layer to the output layer to find $q^{th}$ **order outputs**, $(y_k^l)^q$ and the required derivatives and sensitivities for BP such as $f'(x_k^l)$, $\nabla_y\Psi_{ki}^{l+1}$, $\nabla_{\Psi_{ki}}P_i^{l+1}$ and $\nabla_w\Psi_{ki}^{l+1}$ of each neuron, *k*, at each layer, *l*. |
|      ii. **BP**: Compute delta error at the output layer and then using Eqs. **(14)** and **(16)** back-propagate the error back to the first hidden layer to compute delta errors of each neuron, *k*, $\Delta_k^l$ at each layer, *l*. |
|      iii. **PP**: Find the bias and weight sensitivities using Eqs. **(22)** and **(23)**, respectively. |
|      iv. **Update**: Update the weights and biases with the (cumulation of) sensitivities found in previous step scaled with the learning factor, ε, as in Eq. (49): |
| 3) **Return** Self-*ONN\** |

## B. BP for Non-localized Kernel Operations by Random Bias

In a conventional BP, starting from the output (operational) layer, the error is back-propagated to the 1st hidden layer. For the sake of simplicity, for an image **I** in the training dataset suppose that the error (loss) function is L2-loss or the Mean-Square-Error (MSE) error function, $E(I)$, is used can be expressed as,

$$E(I) = \frac{1}{|I|} \sum_p \left(y_1^L(I_p) - T(I_p)\right)^2 \qquad (25)$$

where $I_p$ is the pixel $p$ of the image $I$, $T$ is the target output and $y_1^L$ is the predicted output. The delta error in the output layer of the input map can then be expressed in Eq. (26).

$$\Delta_1^L = \frac{\partial E}{\partial x_1^L} = \frac{\partial E}{\partial y_1^L}\frac{\partial y_1^L}{\partial x_1^L} = \frac{2}{|I|}(y_1^L(I) - T(I))f'(x_1^L(I)) \qquad (26)$$

For Self-ONNs with super neurons, the contributions of each shifted pixel in the output map, $y_k^l(m + \alpha_k^i, n + \beta_k^i)$, on the next layer input map, $x_i^{l+1}(m,n)$, can now be expressed as in Eq. (27)

(highlighted in red for clarity). So for the hidden operational layers, a close look at Eq. (27) will reveal the fact that the contributions of each pixel in the $(M + 2\Gamma) \times (N + 2\Gamma)$ shifted output map, $y_k^l(m + \alpha_k^i, n + \beta_k^i)$ on the next layer input maps, $x_i^{l+1}(m,n), i \in [1, N_{l+1}]$, depend solely on the bias of each connection, $(\alpha_k^i, \beta_k^i)$. Therefore, the delta error of the output pixel, $y_k^l(m,n)$, should be computed for each connection and then cumulated. Using the chain rule, the delta error of the output pixel, $y_k^l(m,n)$, can therefore, be expressed as in Eq. (28) in the generic form of pool, $P_i^{l+1}$, and composite nodal operator function, $\Psi$, of the $i^{th}$ super neuron, $i \in [1,..,N_{l+1}]$. In Eq. (28), note that the first term, $\frac{\partial x_i^{l+1}(m-r,n-t)}{\partial P_i^{l+1}\left[..,\Psi\left(y_k^l(m+\alpha_k^i,n+\beta_k^i),w_{ik}^{l+1}(r,t)\right),..\right]} = 1$. Let the (shifted) 4D matrices $\nabla_\Psi P_i^{l+1}(m + \alpha_k^i, n + \beta_k^i, r, t) = \frac{\partial P_i^{l+1}\left[..,\Psi\left(y_k^l(m+\alpha_k^i,n+\beta_k^i),w_{ik}^{l+1}(r,t)\right),..\right]}{\partial \Psi\left(y_k^l(m+\alpha_k^i,n+\beta_k^i),w_{ik}^{l+1}(r,t)\right)}$ and $\nabla_y \Psi(m + \alpha_k^i, n + \beta_k^i, r, t) = \frac{\partial \Psi\left(y_k^l(m+\alpha_k^i,n+\beta_k^i),w_{ik}^{l+1}(r,t)\right)}{\partial y_k^l(m+\alpha_k^i,n+\beta_k^i)}$. Then, Eq. (28) simplifies to Eq. (29).

$$\begin{aligned}
x_i^{l+1}(m-1, n-1) &= \ldots + \\
P_i^{l+1}\bigl[\Psi\bigl(y_k^l(m+\alpha_k^i-1, n+\alpha_k^i-1), w_{ik}^{l+1}(\mathbf{0,0})\bigr), \ldots, \Psi\bigl(y_k^l(m+\alpha_k^i, n+\beta_k^i), w_{ik}^{l+1}(\mathbf{1,1})\bigr)\bigr] &+ \ldots \\
x_i^{l+1}(m-1, n) &= \ldots + \\
P_i^{l+1}\bigl[\Psi\bigl(y_k^l(m+\alpha_k^i-1, n+\alpha_k^i), w_{ik}^{l+1}(\mathbf{0,0})\bigr), \ldots, \Psi\bigl(y_k^l(m+\alpha_k^i, n+\beta_k^i), w_{ik}^{l+1}(\mathbf{1,0})\bigr), \ldots\bigr] &+ \ldots \\
x_i^{l+1}(\mathbf{m,n}) &= \ldots + \\
P_i^{l+1}\bigl[\Psi\bigl(y_k^l(m+\alpha_k^i, n+\beta_k^i), w_{ik}^{l+1}(\mathbf{0,0})\bigr), \ldots, \Psi\bigl(y_k^l(m+\alpha_k^i+r, n+\alpha_k^i+t), w_{ik}^{l+1}(\mathbf{r,t})\bigr) \ldots\bigr] &+ \ldots \\
\ldots & \\
\therefore x_i^{l+1}(m-r, n-t)\Big|_{(1,1)}^{(M-1,N-1)} &= b_i^{l+1} + \sum_{k=1}^{N_1} P_i^{l+1}\bigl[\ldots, \Psi\bigl(y_k^l(m+\alpha_k^i, n+\beta_k^i), w_{ik}^{l+1}(r,t)\bigr), \ldots\bigr]
\end{aligned} \qquad (27)$$

$$\begin{aligned}
\therefore \frac{\partial E}{\partial y_k^l}(m + \alpha_k^i, n + \beta_k^i)\bigg|_{(0,0)}^{(M-1,N-1)} &= \Delta y_k^l(m + \alpha_k^i, n + \beta_k^i) = \\
\sum_{r=0}^{K_x-1}\sum_{t=0}^{K_y-1} \frac{\partial E}{\partial x_i^{l+1}(m-r,n-t)} &\times \frac{\partial x_i^{l+1}(m-r,n-t)}{\partial P_i^{l+1}\left[..,\Psi\left(y_k^l(m+\alpha_k^i,n+\beta_k^i),w_{ik}^{l+1}(r,t)\right),..\right]} \times \\
\frac{\partial P_i^{l+1}\left[..,\Psi\left(y_k^l(m+\alpha_k^i,n+\beta_k^i),w_{ik}^{l+1}(r,t)\right),..\right]}{\partial \Psi\left(y_k^l(m+\alpha_k^i,n+\beta_k^i),w_{ik}^{l+1}(r,t)\right)} &\times \frac{\partial \Psi\left(y_k^l(m+\alpha_k^i,n+\beta_k^i),w_{ik}^{l+1}(r,t)\right)}{\partial y_k^l(m+\alpha_k^i,n+\beta_k^i)}
\end{aligned} \qquad (28)$$

$$\Delta y_k^l(m + \alpha_k^i, n + \beta_k^i) = \sum_{r=0}^{K_x-1}\sum_{t=0}^{K_y-1} \Delta_i^{l+1}(m-r, n-t) \times \nabla_\Psi P_i^{l+1}(m + \alpha_k^i, n + \beta_k^i, r, t) \times \nabla_y \Psi(m + \alpha_k^i, n + \beta_k^i, r, t) \qquad (29)$$

where $\nabla_y \Psi(m + \alpha_k^i, n + \beta_k^i, r, t)$ can be directly computed as,

$$\begin{aligned}
\nabla_y \Psi(m + \alpha_k^i, n + \beta_k^i, r, t) &= w_{ik}^{l+1}(r,t,1) + 2w_{ik}^{l+1}(r,t,2)y_k^l(m + \alpha_k^i, n + \beta_k^i) + \cdots \\
&\quad + Q w_{ik}^{l+1}(r,t,Q) y_k^l(m + \alpha_k^i, n + \beta_k^i)^{Q-1}
\end{aligned} \qquad (30)$$

Now let $\nabla_y P_i^{l+1}(m + \alpha_k^i, n + \beta_k^i, r, t) = \nabla_\Psi P_i^{l+1}(m + \alpha_k^i, n + \beta_k^i, r, t) \times \nabla_y \Psi(m + \alpha_k^i, n + \beta_k^i, r, t)$. In this study, the *summation* is used as the pool operator for the sake of simplicity, i.e., $P_i^{l+1} = \Sigma$, so, $\nabla_\Psi P_i^{l+1}(m, n, r, t) = 1$ and thus, $\nabla_y P_i^{l+1}(m + \alpha_k^i, n + \beta_k^i, r, t) = \nabla_y \Psi(m + \alpha_k^i, n + \beta_k^i, r, t)$ then the delta error computed for the connection to the $i^{th}$ neuron at layer $l+1$ can be expressed as,

$$\begin{aligned}
\mathbf{T}^{(\alpha_k^i, \beta_k^i)}(\Delta y_k^l) &= \Delta y_k^l(m + \alpha_k^i, n + \beta_k^i) \\
&= Conv2Dvar\left\{\Delta_i^{l+1}, \mathbf{T}^{(\alpha_k^i, \beta_k^i)}(\nabla_y \Psi)\right\}
\end{aligned} \qquad (31)$$

Finally, the overall delta error for the output map, $\Delta y_k^l$, is computed as the cumulation of the back-shifted individual delta-errors, i.e.,

$$\Delta y_k^l(m,n)\big|_{(0,0)}^{(M-1,N-1)} = \sum_{i=1}^{N_{l+1}} \mathbf{T}^{(-\alpha_k^i,-\beta_k^i)}\left(\Delta y_i^l(m+\alpha_k^i, n+\beta_k^i)\right) \quad (32)$$

Once the $\Delta y_k^l$ is computed, using the chain-rule, one can finalize the back-propagation of the delta error from layer $l+1$ to layer $l$, as follows:

$$\Delta_k^l = \frac{\partial E}{\partial x_k^l} = \frac{\partial E}{\partial y_k^l}\frac{\partial y_k^l}{\partial x_k^l} = \frac{\partial E}{\partial y_k^l}f'(x_k^l) = \Delta y_k^l f'(x_k^l) \quad (33)$$

When there is a pooling (down-sampling) by factors, *ssx*, and *ssy*, then the back-propagated delta-error by Eq. (33) should be first up-sampled to compute the delta-error of the neuron. Let zero-order up-sampled map be: $uy_k^l = \text{up}_{ssx,ssy}(y_k^l)$. Then Eq. (33) can be modified, as follows:

$$\Delta_k^l = \frac{\partial E}{\partial x_k^l} = \frac{\partial E}{\partial y_k^l}\frac{\partial y_k^l}{\partial x_k^l} = \frac{\partial E}{\partial y_k^l}\frac{\partial y_k^l}{\partial uy_k^l}\frac{\partial uy_k^l}{\partial x_k^l} = \text{up}_{ssx,ssy}(\Delta y_k^l)\beta f'(x_k^l) \quad (34)$$

where $\beta = \frac{1}{ssx.ssy}$ since each pixel of $y_k^l$ is now obtained by averaging $(ssx.ssy)$ number of pixels of the intermediate output, $uy_k^l$. Finally, when there is an up-sampling by factors, *usx*, and *usy*, then let the average-pooled map be: $dy_k^l = \text{down}_{usx,usy}(y_k^l)$. Then Eq. (33) can be updated as follows:

$$\Delta_k^l = \frac{\partial E}{\partial x_k^l} = \frac{\partial E}{\partial y_k^l}\frac{\partial y_k^l}{\partial x_k^l} = \frac{\partial E}{\partial y_k^l}\frac{\partial y_k^l}{\partial dy_k^l}\frac{\partial dy_k^l}{\partial x_k^l} = \text{down}_{usx,usy}(\Delta y_k^l)\beta^{-1} f'(x_k^l) \quad (35)$$

As for the computation of the sensitivities for kernel parameters, $\frac{\partial E}{\partial w_{ik}^{l+1}}(r,t,q)\big|_{(0,0,1)}^{(Kx-1,Ky-1,Q)}$, and bias, $\frac{\partial E}{\partial b_k^l}$, Eq. (3) indicates that the $q^{\text{th}}$ element of the array, $\mathbf{w}_{ik}^{l+1}(\mathbf{r},\mathbf{t})$, contributes to all the pixels of $x_i^{l+1}(m,n)$. Once again by using the chain rule of partial derivatives, the sensitivities for kernel parameters can be expressed in Eq. (36). Since $\frac{\partial \Psi(y_k^l(m+\alpha_k^i+r,n+\beta_k^i+t), w_{ik}^{l+1}(r,t))}{\partial w_{ik}^{l+1}(r,t,q)} = y_k^l(m+\alpha_k^i+r, n+\beta_k^i+t)^q$, $P_i^{l+1} = \Sigma$, and $\nabla_\Psi P_i^{l+1}(m+\alpha_k^i+r, n+\beta_k^i+t) = 1$. then Eq. (36) simplifies to Eq. (37).

$$\frac{\partial E}{\partial w_{ik}^{l+1}}(r,t,q)\bigg|_{(0,0,1)}^{(Kx-1,Ky-1,Q)} = \sum_{m=0}^{M-Kx+1}\sum_{n=0}^{N-Ky+1}\left(\begin{array}{c}\frac{\partial E}{\partial x_i^{l+1}(m,n)}\times\frac{\partial x_i^{l+1}(m,n)}{\partial P_i^{l+1}[..,\Psi(y_k^l(m+\alpha_k^i+r,n+\beta_k^i+t), w_{ik}^{l+1}(r,t))...]}\times \\ \frac{\partial P_i^{l+1}[..,\Psi(y_k^l(m+\alpha_k^i+r,n+\beta_k^i+t), w_{ik}^{l+1}(r,t)),...]}{\partial \Psi(y_k^l(m+\alpha_k^i+r,n+\beta_k^i+t), w_{ik}^{l+1}(r,t))}\times \\ \frac{\partial \Psi(y_k^l(m+\alpha_k^i+r,n+\beta_k^i+t), w_{ik}^{l+1}(r,t))}{\partial w_{ik}^{l+1}(r,t,q)}\end{array}\right) \quad (36)$$

$$\frac{\partial E}{\partial w_{ik}^{l+1}}(r,t,q)\bigg|_{(0,0,1)}^{(Kx-1,Ky-1,Q)} = \sum_{m=0}^{M-Kx}\sum_{n=0}^{N-Ky}\Delta_i^{l+1}(m,n)\times\nabla_\Psi P_i^{l+1}(m+\alpha_k^i+r, n+\beta_k^i+t)\times y_k^l(m+\alpha_k^i+r, n+\beta_k^i+t)^q$$
$$= \sum_{m=0}^{M-Kx}\sum_{n=0}^{N-Ky}\Delta_i^{l+1}(m,n)\times y_k^l(m+\alpha_k^i+r, n+\beta_k^i+t)^q \quad (37)$$
$$\therefore \frac{\partial E}{\partial w_{ik}^{l+1}}\langle q\rangle = conv2D\left(\Delta_i^{l+1}, \left(\mathbf{T}^{(\alpha_k^i,\beta_k^i)}(y_k^l)\right)^q, 'NoZeroPad'\right)$$

For the bias sensitivity, the chain rule yields:

$$\frac{\partial E}{\partial b_k^l} = \sum_{m=0}^{M-1}\sum_{n=0}^{N-1}\frac{\partial E}{\partial x_k^l(m,n)}\frac{\partial x_k^l(m,n)}{\partial b_k^l} = \sum_{m=0}^{M-1}\sum_{n=0}^{N-1}\Delta_k^l(m,n) \quad (38)$$

## C. BP for Non-localized Kernel Operations by the BP-optimized Bias

Recall that Eq. (6) allows us to compute the derivatives of the output map w.r.t the individual bias elements, as expressed in Eq. (41). These derivatives will be needed in the BP formulation that will be covered in this section.

The delta error in the output layer of the input map is the same as in Eq. (25). With $\vec{y}_k^l(m, n) = y_k^l(m + \alpha_k^i, n + \beta_k^i)$ Eq. (28) can be simplified as in Eq. (42) and with $P_i^{l+1} = \Sigma$, it yields Eq. (43) where $\nabla_{\vec{y}} P_i^{l+1}(m,n,r,t) = \nabla_{\Psi} P_i^{l+1}(m,n,r,t) \times \nabla_{\vec{y}} \Psi(m,n,r,t) = \nabla_{\vec{y}} \Psi(m,n,r,t)$ and $\nabla_{\vec{y}} \Psi(m,n,r,t)$ can be directly computed as in Eq. (44). Finally, the delta error of $\vec{y}_k^l$ (from its contribution to $x_i^{l+1}$ alone) can be computed as,

$$\mathbf{T}^{(\alpha_k^i, \beta_k^i)}(\Delta y_k^l) = \Delta \vec{y}_k^l = Conv2Dvar\{\Delta_i^{l+1}, (\nabla_{\vec{y}} \Psi)\} \quad (39)$$

Basically, in these equations, we are using the grid of $\vec{y}_k^l(m, n)$ - not the original grid of $y_k^l$. However, we need to compute individual $\Delta y_k^l$ from the $\Delta \vec{y}_k^l$ for each connection in the next layer so that we can cumulate them to compute the overall delta error for $y_k^l$. To accomplish this, as in the earlier approach with random (integer) bias, the overall delta error for the output map, $\Delta y_k^l$, will be computed as the cumulation of the back-shifted individual delta-errors, $\Delta \vec{y}_k^l$ computed for each connection, i.e.,

$$\Delta y_k^l(m,n)\Big|_{(0,0)}^{(M-1,N-1)} = \sum_{i=1}^{N_{l+1}} \mathbf{T}^{(-\alpha_k^i, -\beta_k^i)}\left(\Delta y_k^l(m + \alpha_k^i, n + \beta_k^i)\right)$$

$$= \sum_{i=1}^{N_{l+1}} \mathbf{T}^{(-\lfloor\alpha_k^i\rfloor, -\lfloor\beta_k^i\rfloor)}\left(\Delta y_k^l(m_\alpha, n_\beta)\right) \quad (40)$$

where $m_\alpha = m + \lfloor\alpha_k^i\rfloor$ and $n_\beta = n + \lfloor\beta_k^i\rfloor$. Since the bias elements are not an integer, we should now use the reverse-interpolation to compute first, $\Delta y_k^l(m + \lfloor\alpha_k^i\rfloor, n + \lfloor\beta_k^i\rfloor)$ as illustrated in Figure 11. Once again using bilinear interpolation, $\Delta y_k^l(m_\alpha, n_\beta)$ can be computed as expressed in Eq. (45). As in the random bias approach, the overall delta error for the output map, $\Delta y_k^l$, is computed as the cumulation of the back-shifted individual delta-errors using Eq. (40). Once on the integer grid, it is straightforward to compute $\Delta y_k^l$ using Eq. (32).

After the (overall) $\Delta y_k^l$ is computed, using Eq. (33) (or Eq. (34) or (35) in case down- or up-sampling is performed), the delta error, $\Delta_k^l$, can be computed and hence, the back-propagation of the (delta) error from layer $l+1$ to the $k^{th}$ neuron at layer $l$ is completed.

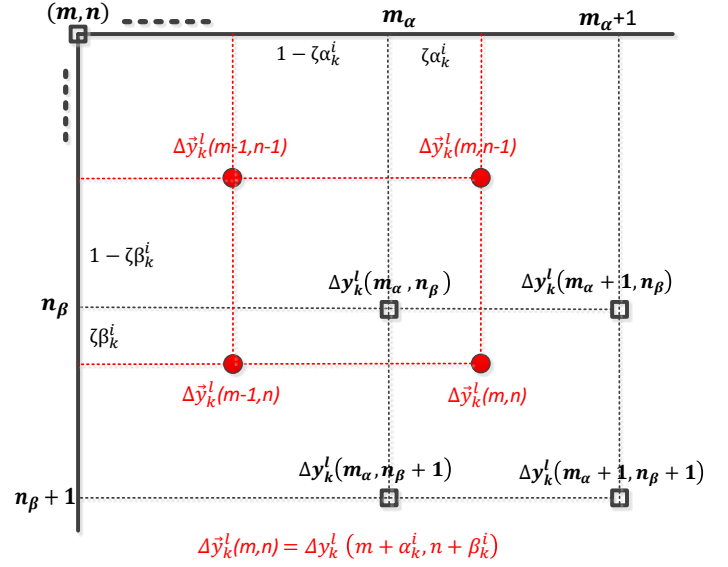

**Figure 11: The reverse interpolation from the shifted delta error, $\Delta \vec{y}_k^l(m, n) = \Delta y_k^l(m + \alpha_k^i, n + \beta_k^i)$ by the bias, $(\alpha_k^i, \beta_k^i) \in \mathbb{R}$, to the original delta error with integer shifts, $\Delta y_k^l(m + \lfloor\alpha_k^i\rfloor, n + \lfloor\beta_k^i\rfloor)$ where $(\lfloor\alpha_k^i\rfloor, \lfloor\beta_k^i\rfloor) \in \mathbb{Z}$.**

Once the back-propagation of delta errors is completed, then weight and bias sensitivities can be computed using Eqs. (37) and (38) with the same simplifications. Note that $\vec{y}_k^l = \mathbf{T}^{(\alpha_k^i, \beta_k^i)}(y_k^l)$ is the shifted (interpolated) output map as before with the only difference that $(\alpha_k^i, \beta_k^i) \in \mathbb{R}$.

Finally, for the spatial bias sensitivities, $\Delta\alpha_k^i = \frac{\partial E}{\partial \alpha_k^i}, \Delta\beta_k^i = \frac{\partial E}{\partial \beta_k^i}$, the spatial bias pair, $(\alpha_k^i, \beta_k^i)$, shifts only the pixels of the output map, $y_k^l$, to contribute to all pixels of $x_i^{l+1}$. By using the chain rule of partial derivatives, the sensitivities of the spatial bias pair can be expressed in Eq. (46). Let $\nabla_\alpha \vec{y}(m,n) = \frac{\partial \vec{y}_k^l(m,n)}{\partial \alpha_k^i}$, which was expressed in Eq. (41), Eq. (46) finally simplifies to Eq. (47) where $\Delta\alpha_k^i$ is a scalar and $\nabla_\alpha \vec{y} \otimes \Delta\vec{y}_k^l$ is the 2D cross-correlation between $\nabla_\alpha \vec{y}$ and $\Delta \vec{y}_k^l$.

$$\frac{\partial \vec{y}_k^l(m+r, n+t)}{\partial \alpha_k^i} = (1 - \zeta\beta_k^i)\left(y_k^l(m_\alpha^r + 1, n_\beta^t) - y_k^l(m_\alpha^r, n_\beta^t)\right) +$$
$$\zeta\beta_k^i\left(y_k^l(m_\alpha^r + 1, n_\beta^t + 1) - y_k^l(m_\alpha^r, n_\beta^t + 1)\right)$$

$$\frac{\partial \vec{y}_k^l(m+r, n+t)}{\partial \beta_k^i} = (1 - \zeta\alpha_k^i)\left(y_k^l(m_\alpha^r, n_\beta^t + 1) - y_k^l(m_\alpha^r, n_\beta^t)\right) +$$
$$\zeta\alpha_k^i\left(y_k^l(m_\alpha^r + 1, n_\beta^t + 1) - y_k^l(m_\alpha^r + 1, n_\beta^t)\right)$$

(41)

$$\left.\frac{\partial E}{\partial y_k^l}\left(m+\alpha_k^i, n+\beta_k^i\right)\right|_{(0,0)}^{(M-1,N-1)} = \Delta y_k^l\left(m+\alpha_k^i, n+\beta_k^i\right) = \Delta \vec{y}_k^l(m,n)$$

$$= \sum_{r=0}^{K_x-1} \sum_{t=0}^{K_y-1} \frac{\frac{\partial E}{\partial x_i^{l+1}(m-r,n-t)} \times \frac{\partial x_i^{l+1}(m-r,n-t)}{\partial P_i^{l+1}[..,\Psi(\vec{y}_k^l(m,n),w_{ik}^{l+1}(r,t)),..]} \times}{\frac{\partial P_i^{l+1}[..,\Psi(\vec{y}_k^l(m,n),w_{ik}^{l+1}(r,t)),..]}{\partial \Psi(\vec{y}_k^l(m,n),w_{ik}^{l+1}(r,t))} \cdot \frac{\partial \Psi(\vec{y}_k^l(m,n),w_{ik}^{l+1}(r,t))}{\partial \vec{y}_k^l(m,n)}}$$ (42)

$$\Delta y_k^l(m+\alpha_k^i, n+\beta_k^i) = \Delta \vec{y}_k^l(m,n) = \sum_{r=0}^{K_x-1} \sum_{t=0}^{K_y-1} \Delta_i^{l+1}(m-r,n-t) \times \nabla_{\vec{y}} \Psi(m,n,r,t) \quad (43)$$

$$\nabla_{\vec{y}} \Psi(m,n,r,t) = w_{ik}^{l+1}(r,t,1) + 2w_{ik}^{l+1}(r,t,2)\vec{y}_k^l(m,n) + \cdots + Q w_{ik}^{l+1}(r,t,Q)\vec{y}_k^l(m,n)^{Q-1} \quad (44)$$

$$\Delta y_k^l(m_\alpha^r, n_\beta^t) = \Delta \vec{y}_k^l(m,n)(1-\zeta\alpha_k^i)(1-\zeta\beta_k^i) + \Delta \vec{y}_k^l(m-1, n-1)\zeta\alpha_k^i \zeta \beta_k^i + \\ \Delta \vec{y}_k^l(m-1, n)\zeta\alpha_k^i(1-\zeta\beta_k^i) + \Delta \vec{y}_k^l(m, n-1)(1-\zeta\alpha_k^i)\zeta\beta_k^i \quad (45)$$

$$\frac{\partial E}{\partial \alpha_k^i} = \Delta \alpha_k^i =$$

$$\sum_{m=0}^{M-1} \sum_{n=0}^{N-1} \left( \sum_{r=0}^{K_x-1} \sum_{t=0}^{K_y-1} \frac{\frac{\partial E}{\partial x_i^{l+1}(m-r,n-t)} \times \frac{\partial x_i^{l+1}(m-r,n-t)}{\partial P_i^{l+1}[..,\Psi(\vec{y}_k^l(m,n),w_{ik}^{l+1}(r,t)),..]} \times}{\frac{\partial P_i^{l+1}[..,\Psi(\vec{y}_k^l(m,n),w_{ik}^{l+1}(r,t)),..]}{\partial \Psi(\vec{y}_k^l(m,n),w_{ik}^{l+1}(r,t))} \cdot \frac{\partial \Psi(\vec{y}_k^l(m,n),w_{ik}^{l+1}(r,t))}{\partial \vec{y}_k^l(m,n)} \times \frac{\partial \vec{y}_k^l(m,n)}{\partial \alpha_k^i}} \right) \quad (46)$$

$$\Delta \alpha_k^i = \frac{\partial E}{\partial \alpha_k^i} = \sum_{m=0}^{M-1} \sum_{n=0}^{N-1} \nabla_\alpha \mathbf{y}(m,n) \times$$
$$\left( \sum_{r=0}^{K_x-1} \sum_{t=0}^{K_y-1} \Delta_i^{l+1}(m-r,n-t) \times \nabla_\Psi P_i^{l+1}(m+\alpha_k^i, n+\beta_k^i, r, t) \times \nabla_{\vec{y}} \Psi(m+\alpha_k^i, n+\beta_k^i, r, t) \right) \quad (47)$$
$$= \sum_{m=0}^{M-1} \sum_{n=0}^{N-1} \nabla_\alpha \vec{y}(m,n) \times \Delta \vec{y}_k^l(m,n) = \nabla_\alpha \vec{y} \otimes \Delta \vec{y}_k^l$$

Similarly, it is straightforward to show that the sensitivity, $\Delta \beta_k^i = \frac{\partial E}{\partial \beta_k^i}$, can be expressed as,

$$\Delta \beta_k^i = \frac{\partial E}{\partial \beta_k^i} = \sum_{m=0}^{M-1} \sum_{n=0}^{N-1} \nabla_\beta \vec{y}(m,n) \times \Delta \vec{y}_k^l(m,n) \\ = \nabla_\beta \vec{y} \otimes \Delta \vec{y}_k^l \quad (48)$$

where $\nabla_\beta \vec{y}(m,n) = \frac{\partial \vec{y}_k^l(m,n)}{\partial \beta_k^i}$ as expressed in Eq. (41). It is interesting to see that both spatial bias sensitivities depend on the cross-correlation of two distinct gradients, the shifted (interpolated) output map delta error and its direct derivative w.r.t the corresponding bias element. This means that during BP iterations, the ongoing gradient descent operation, e.g. Stochastic Gradient Descent (SGD), will keep updating the kernel location until either correlation between these two gradients vanishes (e.g., they become uncorrelated) or when the (magnitude of the) delta errors diminishes eventually at the final stages of the BP (e.g. convergence of the gradient descent). In other words, the local optimal location of a particular kernel of a particular connection -if exists for the particular problem at hand- will be converged when either of the conditions is satisfied (i.e., when $\Delta \alpha_k^i, \Delta \beta_k^i \approx 0$).

During each BP iteration, $t$, the kernel parameters, $w_{ik}^{l+1}\langle q \rangle(t)$, and biases, $b_i^l(t)$, (spatial) bias pairs, $\alpha_k^i(t), \beta_k^i(t)$, of each super neuron in the Self-ONN are updated until a stopping criterion is met. Let, $\varepsilon(t)$ and $\gamma(t)$ be the learning factors at iteration, $t$, of weights and spatial bias pairs, respectively. One can express the SGD update for the kernel parameters, bias, and the kernel location of each super neuron, $i$, at the layer, $l$, in Eq. (49). The parameters of a Self-ONN for BP training via SGD are presented in Table 4.

$$\begin{aligned}
w_{ik}^l\langle q\rangle(t+1) &= w_{ik}^l\langle q\rangle(t) - \varepsilon(t)\frac{\partial E}{\partial w_{ik}^l}\langle q\rangle, & q &\in [1,Q], i \in [1,N_l], k \in [1,N_{l-1}] \\
b_i^l(t+1) &= b_i^l(t) - \varepsilon(t)\frac{\partial E}{\partial b_i^l}, & i &\in [1,N_l] \\
\alpha_k^i(t+1) &= \alpha_k^i(t) - \gamma(t)\Delta\alpha_k^i, & i &\in [1,N_l], k \in [1,N_{l-1}] \\
\beta_k^i(t+1) &= \beta_k^i(t) - \gamma(t)\Delta\beta_k^i, & i &\in [1,N_l], k \in [1,N_{l-1}]
\end{aligned} \qquad (49)$$

**Table 4: The train parameters of a Self-ONN with super neurons.**

| Parameter | Symbol | Default | Description |
|---|---|---|---|
| **Max. no. of iterations** | $maxIter$ | 200 | BP stopping criterion |
| **Target min. MSE** | $minMSE$ | $10^{-3}$ | BP stopping criterion |
| **Range for rand. initialization** | $U(-\varpi, \varpi)$ | $\varpi = 0.1$ | for kernel parameters |
| | $U(-\Gamma, \Gamma)$ | $\Gamma = 8$ | for spatial bias |
| **Learning factors** | $\varepsilon$ | $\varepsilon(t) = 0.1$ | for kernel parameters |
| | $\gamma$ | $\gamma(t) = 10$ | for spatial bias pairs |

**Algorithm 1: Back-Propagation by SGD for Self-ONNs with Super neurons**

**Input**: $Self\text{-}ONN(0)$, $Hyper\ Parameters$, $Train\ Parameters$: $Stopping\ Criteria\ (maxIter, minMSE), \varpi, \Gamma, \varepsilon, \gamma$
**Output**: $Self\text{-}ONN^* = BP\ (Self\text{-}ONN(0), SGD, Hyper\ Parameters, Train\ Parameters)$

1) **Initialize** network parameters of each super neuron:
   a. $w_{ik}^{l+1}\langle q\rangle(0) = U(-\varpi, \varpi), b_i^l(t+1) = U(-\varpi, \varpi)\ for\ \forall i \in [1, N_{l+1}], \forall k \in [1, N_l], \forall q \in [1, Q]$
   b. $\alpha_k^i(0) = U(-\Gamma, \Gamma), \beta_k^i(0) = U(-\Gamma, \Gamma)\ for\ \forall i \in [1, N_{l+1}], \forall k \in [1, N_l]$

2) **UNTIL** either stopping criterion is reached, **ITERATE** ($t = 1: maxIter$):
   a. **For** each batch in the train dataset, **DO**:
      i. **Init**: Assign next item, $I_p$, directly as the output map(s) in the input layer neurons and using Eq. (6) create the shifted output map(s) along with their powers, $(\vec{y}_k^0)^q$, $\forall q \in [1, Q_1]$ where $Q_1$ is the polynomial order of the super neurons in the 1st hidden layer.
      ii. **FP**: From the previous layer (shifted) output maps, compute each input map in the 1st hidden layer, $x_i^1$, $\forall i \in [1, N_1]$ using Eq. (7), then the native output maps, $y_i^1$ and finally, the shifted output maps along with their powers, $(\vec{y}_i^1)^q$ $\forall q \in [1, Q_1]$.
      iii. **FP**: Then compute the required derivatives and sensitivities for each hidden layer, such as $f'(x_k^l)$, $\nabla_y \Psi_{ik}^l$, and $\nabla_w \Psi_{ik}^l$ of each neuron, $i$, at each layer, $l$. ($\nabla_{\Psi_{ik}} P_i^l = 1$)
      iv. **FP**: Repeat (ii) until the output layer is reached. Compute the output map(s), $y_1^L(I_p)$, of the neurons in the output layer and then, compute the MSE and delta error, $\Delta_1^L$, using Eqs. (25) and (26), respectively.
      v. **BP**: For each hidden neuron at the last hidden layer, using Eq. (39) compute delta error for the shifted output map and then using Eq. (45), perform reverse-interpolation (and shift) to compute the delta error of the actual output map for each connection to the next layer.
      vi. **BP**: Using Eq. (32) compute the overall delta error for the output map, $\Delta y_k^l$, as the cumulation of the back-shifted individual delta errors.
      vii. **BP**: Finally, using Eq. (33) (or Eq. (34) or (35) in case down- or up-sampling is performed), compute the delta error at this level, $\Delta_k^l$.
      viii. **PP**: Compute sensitivities for the kernel parameters, bias, and spatial bias pair using Eqs. (37), (38), (47), and (48) respectively.
      ix. **Update**: Update for the kernel parameters, bias, and the kernel location of each super neuron in the network with the (cumulation of) sensitivities found in step (viii) scaled with the current learning factors, $\varepsilon(t)$ and $\gamma(t)$, using Eq. (49).

3) **Return Self-ONN\***

To initiate the BP training by SGD over a dataset, a Self-ONN is first configured according to the network parameters, i.e., number of layers ($L$) and hidden neurons ($N_l$), the kernel-size ($Kx, Ky$), the pooling type and the (polynomial) order for each layer/neuron are set in advance. Let $Self\text{-}ONN(0)$ be the initially configured network ready for BP training. In the pseudo-code for BP training presented in Alg. 1, five consecutive stages in an iterative loop are visible: 1) BP initialization (Step 1), 2) Forward-Propagation (FP) of each image in the batch where native and shifted (interpolated) output maps, derivatives and output MSE and delta error are computed (in Step 2.a, i –iii), 3) Back-Propagation (BP) of the delta error from the output layer to the first hidden layer (in Step 2.a, v –vii), 4) post-processing (PP) where the kernel parameter and bias sensitivities, the sensitivities of the spatial bias pair are computed for each image in the batch and cumulated, and 5) Update: when all images in the batch are processed, then the kernel, bias and the kernel location of each super neuron in the network are updated and this is repeated for the other batches and iterations. The pseudo-code In Alg. 1 can be used for a Self-ONN with super neurons that are configured with the non-localized kernel operations by random spatial bias, the following steps should be modified accordingly. First, the initialization of bias elements should be an integer in 1.b., i.e., $\alpha_k^i(0) = \lfloor U(-\mathbf{\Gamma} - \mathbf{1}, \mathbf{\Gamma} + \mathbf{1}) \rfloor$ and $\beta_k^i(0) = \lfloor U(-\mathbf{\Gamma} - \mathbf{1}, \mathbf{\Gamma} + \mathbf{1}) \rfloor$ for $\forall i \in [1, N_{l+1}], \forall k \in [1, N_l]$. Then since the spatial bias elements are integers now, Eq. (3) can be used instead of Eq. (7) for FP. Steps 2.a.iv and 2.a.vii are identical for both approaches. The main difference in BP is step 2.a.v where Eq. (31) should be used instead of Eq. (39) for the delta error computed for the connection to the $i^{th}$ neuron at layer $l$+1 and there is no need for reverse interpolation, hence Eq. (45) is simply omitted. Obviously for post-processing (PP) at step 2.a.vii, and Update at step 2.a.xi, Eqs. (47), (48), and (49) are, too, omitted since there is no gradient computation for the spatial bias pair, $\alpha_k^i$, and $\beta_k^i$, as they are fixed as integers during step 1. Since the spatial bias elements are integers now, Eq. (3) can be used instead of Eq. (7) for FP. Steps 2.a.iv and 2.a.vii are identical for both approaches. The main difference in BP is step 2.a.v where Eq. (31) should be used instead of Eq. (39) for the delta error computed for the connection to the $i^{th}$ neuron at layer $l$+1 and there is no need for reverse interpolation, hence Eq. (45) is simply omitted. Obviously for post-processing (PP) at step 2.a.vii, and update at step 2.a.xi, Eqs. (47), (48), and (49) are, too, omitted since there is no gradient computation for the spatial bias pair, $\alpha_k^i$, and $\beta_k^i$, as they are fixed as integers.

## D. Proof of Concept

In order to validate the super neurons' ability to learn the true shift using BP-optimization of the spatial bias pair, a Self-ONN network with one hidden layer and a single neuron is trained over a toy problem where the network aims to learn to regress (transform) an input image to an output image, which is the shifted version of the input image by $(\alpha, \beta) \in \mathbb{Z}[-\Gamma, \Gamma]$, i.e., $y_0^2(m, n) = y_0^0(m + \alpha, n + \beta)$. Therefore, with this setup, we can now validate whether the super neurons with the non-localized kernels are able to learn the true shift collectively during the BP training, and if so, whether the Self-ONN is able to generate the target (shifted) image perfectly well. Figure 12 illustrates this over a sample image where the output image is the shifted version of the input image with $(\alpha = 6, \beta = -7)$ pixels. In this ideal regression case, the cumulative bias shift of the two super neurons in x- and y-directions indeed is equal to the target shift, i.e., $\sum(\alpha_0^0, \beta_0^0) = (6, -7)$ where the 1st order learned kernels are impulses, i.e., $w_{00}^1(r, t) = w_{00}^2(r, t) = \delta(r, t)$. Since this is a validation experiment where the cumulative bias convergence is compared against the actual shift, we keep $Q=1$ to avoid the higher-order (nonlinear) operations and thus to achieve a perfect reconstruction by linear convolution.

The ideal regression case illustrated in Figure 12 shows the configuration of *only* one of the possible BP-optimized super neurons in a Self-ONN. Another ideal output can also be achieved, for instance, when $\sum(\alpha_0^0, \beta_0^0) = (5, -8)$ and the 1st order learned kernels are impulses are $w_{00}^1(r, t) = \delta(r, t)$, $w_{00}^2(r, t) = \delta(r - 1, t - 1)$ *or* $w_{00}^1(r, t) = \delta(r - 1, t - 1)$, $w_{00}^2(r, t) = \delta(r, t)$, *or even*, $w_{00}^1(r, t) = \delta(r - 1, t)$, $w_{00}^2(r, t) = \delta(r, t - 1)$. In this case, the cumulative bias shifts are converged to the close vicinity of the actual shift (with an offset of (1,1) pixels) while the kernels of the hidden and output super neurons with the shifted impulses accommodate for the offset left out by the biases.

Over the 40 input images randomly selected in the Pascal dataset, we created the target images with random shifts by $\Gamma = 8$ pixels. Figure 13 shows four examples of this verification experiment where the input, output, and target images are shown in the first and the last two columns, respectively. The 2nd and 3rd columns show bar plots of the kernels and the 4th column shows the plots of the cumulative bias elements (hidden and output super neurons) in each BP iteration with a blue point. The cumulative, $\sum(\alpha_0^0, \beta_0^0)$, and target shifts, $(\alpha, \beta)$, are shown with the red circles on the plot. The spatial bias pair is initially set as, $(\alpha_0^0, \beta_0^0) = (0,0)$. The BP iterations are stopped when the regression SNR reaches 35dB. In all experiments including the four shown in the figure, the cumulative bias converged to the close vicinity of the actual shift and we observed that offsets such as $(0,1), (1,0)$ or $(1,1)$ pixelsare accommodated by the 2x2 kernels with shifted impulses. This is also visible in the figure where the offset is $(1,1)$ pixels. In the experiments shown in the first and third rows, the kernel functions in the 1st and 2nd (output) layers are: $w_{00}^1(r, t) \cong \delta(r - 1, t - 1)$ and $w_{00}^2(r, t) \cong \delta(r, t)$ while the one in the fourth row, they are: $w_{00}^1(r, t) \cong \delta(r, t - 1)$ and $w_{00}^2(r, t) \cong \delta(r - 1, t)$. Since the early stopping criterion is set as SNR=35dB, the kernels are only approximating the (shifted) impulses. A common observation in all experiments is that the spatial bias elements usually converged during the early stages of the BP, i.e., within around 20-50 iterations while the optimization of the kernels was initiated afterwards.

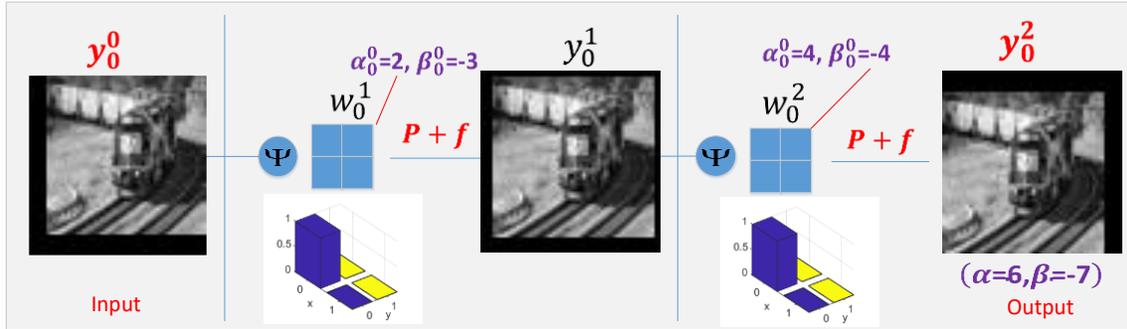

**Figure 12: A sample Self-ONN with a single (hidden) super neuron over the toy problem. The perfect regression of the target is illustrated (SNR = ∞) for an ideal case.**

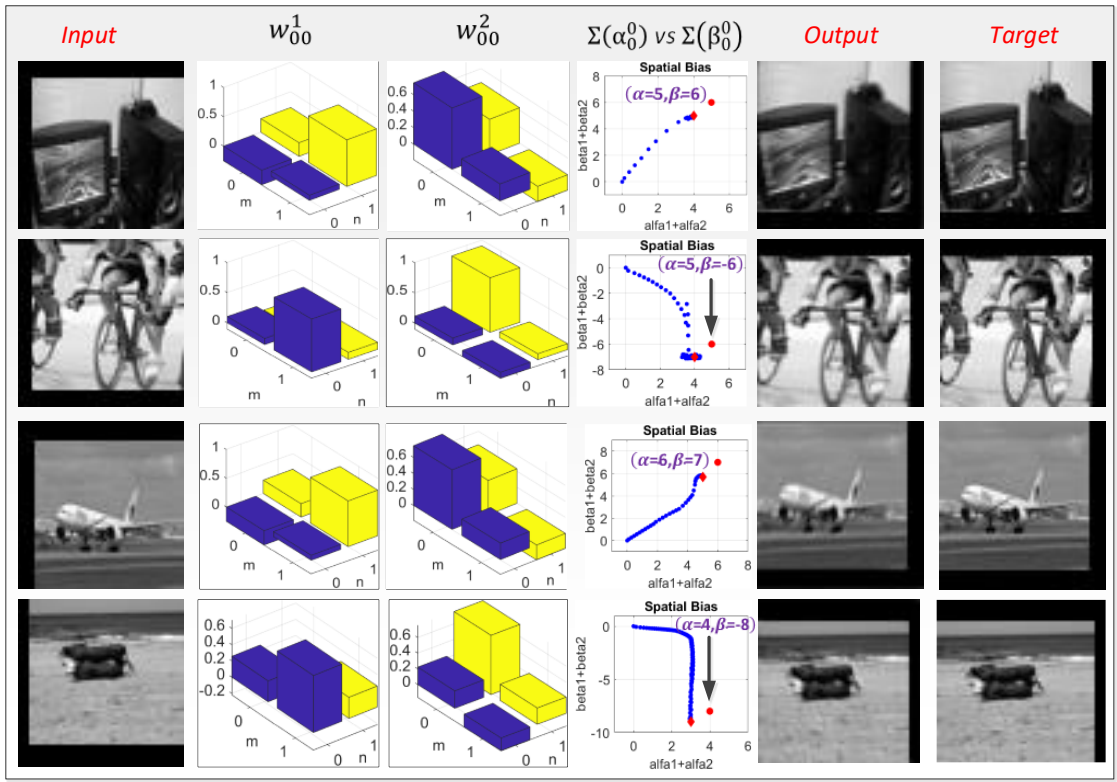

**Figure 13: Four "Proof of Concept" verification experiments where the target images are created with random shifts are shown at each row. The $2^{nd}$ and $3^{rd}$ columns show bar plots of the kernels and the $4^{th}$ column shows the plots of the cumulative bias elements (hidden and output super neurons) in each BP iteration with a blue point. The cumulative, $\sum(\alpha_0^0, \beta_0^0)$, and target shifts $(\alpha, \beta)$ are shown with the red circles on the plot. The BP is stopped at the iteration when the SNR is reached to 35dB.**

In brief, such a "Proof of Concept" demonstration shows a unique capability of the super neurons in a regression problem, i.e., only with a single hidden neuron, from an arbitrary input image, the network can perfectly regress the output image which is the shifted version of the input image. Such an image transformation is not possible for any conventional CNNs, or even Self-ONNs with generative neurons, unless the effective receptive field is expanded by using sufficiently deep and complex networks.